\documentclass[twoside,11pt]{article}

\usepackage{jmlr2e}

\newcommand{\citeg}[1]{\citep[e.g.][]{#1}}
\usepackage{amsmath}
\usepackage{bbm}
\usepackage{url}
\usepackage{booktabs}
\usepackage{microtype}
\usepackage[usenames,dvipsnames]{color}
\usepackage[normalem]{ulem}

\jmlrheading{1}{2000}{1-48}{4/00}{10/00}{
A Framework for Evaluating Approximation Methods for Gaussian Process Regression}

\ShortHeadings{Evaluating Approximation Methods for GPR}
{Chalupka, Williams and Murray}
\firstpageno{1}

\newcommand{\matlab}{{\sc Matlab}}
\def\cut#1{}

\newcommand{\N}{{\mathcal N}}
\newcommand{\defeq}{{\stackrel{\mathit{def}}{=}}}

\newcommand{\balpha}{\mbox{\boldmath{$\alpha$}}}

\newcommand{\bk}{\mathbf{k}}
\newcommand{\bu}{\mathbf{u}}
\newcommand{\bx}{\mathbf{x}}
\newcommand{\by}{\mathbf{y}}

\definecolor{gold}{rgb}{0.85,.66,0}

\newcommand{\te}{\!=\!}
\newcommand{\g}{\!\mid\!}

\newcommand{\tp}{\!+\!}

\begin{document}

\title{A Framework for Evaluating Approximation Methods for Gaussian Process
Regression}

\author{\name Krzysztof Chalupka\setcounter{footnote}{0}\footnotemark \email kjchalup@caltech.edu \\
        \addr Computation and Neural Systems\\
        California Institute of Technology\\
   	1200 E. California Boulevard\\
	Pasadena, CA 91125, USA
	\AND
       \name  Christopher K. I. Williams \email c.k.i.williams@ed.ac.uk \\
       \name Iain Murray \email i.murray@ed.ac.uk \\
       \addr School of Informatics \\
       University of Edinburgh \\
       10 Crichton St, Edinburgh EH8 9AB, UK}
\setcounter{footnote}{1}
\renewcommand{\thefootnote}{\fnsymbol{footnote}}
\footnotetext{This research was carried out when KC was a student at
  the University of Edinburgh.}
\renewcommand{\thefootnote}{\arabic{footnote}}

\maketitle

\begin{abstract}%
  Gaussian process (GP) predictors are an important component of many
  Bayesian approaches to machine learning. However, even a
  straightforward implementation of Gaussian process regression (GPR)
  requires $O(n^2)$ space and $O(n^3)$ time for a dataset of $n$
  examples. Several approximation methods have been proposed, but
  there is a lack of understanding of the relative merits of the
  different approximations, and in what situations they are most
  useful.  We recommend assessing the quality of the predictions
  obtained as a function of the compute time taken, and comparing to
  standard baselines (e.g., Subset of Data and FITC)\@.
  We empirically investigate four different approximation algorithms on
  four different prediction problems, and make our code available to
  encourage future comparisons.
\end{abstract}

\begin{keywords}
Gaussian process regression, subset of data, FITC, local GP\@.
\end{keywords}

\vspace*{0.5cm}

Gaussian process (GP) predictors are widely used in non-parametric
Bayesian approaches to supervised learning problems
\citep{rasmussen-williams-06}.
They can also be used as components for other tasks
including unsupervised learning \citep{lawrence2004}, and
dependent processes for a variety of applications (e.g., \citealt{sudderth2009,adams2010}).
The basic model on which these are based is Gaussian process regression (GPR),
for which a standard implementation
requires $O(n^2)$ space and $O(n^3)$ time for a dataset of $n$
examples, see e.g.\ \citet[ch.~2]{rasmussen-williams-06}.  Several
approximation methods have now been proposed, as detailed
below. Typically the approximation methods are compared to the basic
GPR algorithm. However, as there are now a range of different
approximations, the user is faced with the problem of understanding
their relative merits, and in what situations they are most
useful.

Most approximation algorithms have a tunable complexity parameter,
which we denote as $m$. Our key recommendation is to study the
quality of the predictions obtained as a function of the \emph{compute
time} taken as $m$ is varied, as times can be compared across
different methods. New approximation
methods should be compared against current baselines
like Subset of Data and FITC~(described in Secs.~\ref{sec:sod}--\ref{sec:fitc}).
The time decomposes into that needed for
training the predictor (including setting hyperparameters),
and test time; the user needs to understand which will
dominate in their application. We illustrate this process
by studying four different approximation algorithms on
four different prediction problems. We have published our
code in order to encourage comparisons of other methods
against these baselines.

The structure of the paper is as follows: In Section~\ref{sec:theory} we outline
the complexity of the full GP algorithm and various approximations, and give
some specific details needed to apply them in practice.
Section~\ref{sec:compare} outlines issues that should be considered when
selecting or developing a GP approximation algorithm. Section~\ref{sec:expts}
describes the experimental setup for comparisons, and the results of these
experiments. We conclude with future directions and a discussion.

\section{Approximation algorithms for Gaussian Process Regression (GPR)
\label{sec:theory}}
A regression task has a training set ${\cal D}\te\{\bx_i,y_i\}_{i=1}^n$
with
$D$-dimensional
inputs $\bx_i$ and scalar outputs $y_i$. Assuming that
the outputs are noisy observations of a latent function $f$ at values $f_i\te
f(\bx_i)$, the goal is to compute a predictive distribution over the latent
function value $f_*$
at a test location $\bx_*$.

Assuming a Gaussian process prior over functions $f$ with
zero mean, and covariance or kernel function $k(\cdot, \cdot)$, and
Gaussian observations, $y_i = f_i + \epsilon_i$ where
$\epsilon_i\sim\N(0,\sigma^2)$, gives Gaussian predictions $p(f_*\g\bx_*,{\cal
D})\te\N(\overline{f}_*,\mathbbm{V}[f_*])$, with
predictive mean and variance \citep[see e.g.,][Sec.~2.2]{rasmussen-williams-06}:
\begin{eqnarray}
\overline{f}_* &=& \bk^\top(\bx_*) (K + \sigma^2 I)^{-1} \by \;\;\defeq\;\;
\bk^\top(\bx_*) \balpha, \label{eq:meanpred}\\
\mathbbm{V}[f_*] &=& k(\bx_*,\bx_*) - \bk^\top(\bx_*) (K + \sigma^2 I)^{-1}
\bk(\bx_*), \label{eq:varpred}
\end{eqnarray}
where $K$ is the $n \times n$ matrix with $K_{ij} =
k(\bx_i,\bx_j)$, $\bk(\bx_*)$ is the $n \times 1$ column vector
with the $i$th entry being $k(\bx_*,\bx_i)$, $\by$ is the
column vector of the $n$ target values, and $\balpha =
(K + \sigma^2 I)^{-1} \by$.

The log marginal likelihood of the GPR model is also available in closed form:
\begin{equation}
L = \log p(\by|X) = -\tfrac{1}{2}\, \by^\top (K+\sigma^2_n I)^{-1}\by
-\tfrac{1}{2}\log|K+\sigma^2 I|-\tfrac{n}{2}\log 2\pi.
\label{eq:logmarglike}
\end{equation}
Typically $L$ is viewed as a function
of a set of parameters $\theta$ that specify the kernel. Below we
assume that $\theta$ is set by numerically maximizing $L$ with
a routine like conjugate gradients.  Computation
of $L$ and the gradient $\nabla_{\theta} L$ can be carried out in $O(n^3)$.
Optimizing $L$ is a maximum-likelihood type II or ML-II procedure for $\theta$;
alternatively one might sample over $p(\theta|{\cal D})$
using e.g.\ MCMC\@.
Equations~\ref{eq:meanpred}--\ref{eq:logmarglike} form the
basis of GPR prediction.

We identify three computational phases in carrying out GPR:
\begin{description}
\item [hyperparameter learning:] The hyperparameters are learned, by for
  example maximizing the log marginal likelihood. This is often the
most computationally expensive phase.
\item [training:] Given the hyperparameters,
   all
   computations that don't involve test
  inputs are performed, such as computing $\balpha$ above, and/or
  computing the Cholesky decomposition of $K + \sigma^2_n I$.
  This phase was called ``precomputation'' by
  \citet[Sec.~9.6]{candela2007}.
\item [testing:] Only the computations involving the test inputs are
carried out, those which could not have been done previously. This phase
may be significant if there is a very large test set, or if deploying a trained
model on a machine with limited resources.
\end{description}
Table~\ref{t:approxgpr} lists the computational complexity of
  training and testing full GPR as a function of~$n$. Evaluating the
  marginal likelihood $L$ and its gradient takes more operations than
  `training' (i.e.\ computing the parts of \eqref{eq:meanpred}
  and~\eqref{eq:varpred} that don't depend on $\bx_*$), but has the
  same scaling with~$n$. Hyperparameter learning involves evaluating
  $L$ for all values of the hyperparameters~$\theta$ that are searched
  over, and so is more expensive than training for fixed
  hyperparameters.

\begin{table}[t]
\begin{center}
\begin{tabular}{ccccc} \toprule
Method & Storage& Training & Mean & Variance \\ \midrule
Full & $O(n^2)$ & $O(n^3)$ & $O(n)$ & $O(n^2)$ \\
SoD & $O(m^2)$ & $O(m^3)$ & $O(m)$ & $O(m^2)$ \\
FITC & $O(mn)$ & $O(m^2n)$ & $O(m)$ & $O(m^2)$ \\
Local & $O(mn)$ & $O(m^2n)$ & $O(m)$  & $O(m^2)$ \\ \bottomrule
\end{tabular}
\end{center}
\vspace*{-0.2cm}
\caption{A comparison of the space and time complexity of the
Full, SoD, FITC and Local methods, ignoring the time taken to select the
$m$ subset/inducing points/clusters from the $n$ datapoints.
Training: the time required for preliminary
computations before the test point $\bx_*$ is known, for each hyperparameter setting considered.
Mean (resp.\ variance): the
time needed to compute the predictive mean (variance) at test point~$\bx_*$.
\label{t:approxgpr}}
\end{table}

These complexities can be reduced in special cases, e.g.\
for stationary covariance functions and grid designs, as may be found e.g.\
in geoscience problems. In this case the eigenvectors of $K$ are the
Fourier basis, and matrix inversions etc can be computed analytically.
See e.g.\ \citet{wikle2001,paciorek2007,fritz-neuweiler-nowak-09}
for more details.

Common methods for approximate GPR include Subset of Data (SoD),
  where data points are simply thrown away; inducing point methods
  \citep{candela2005}, where $K$ is approximated by a low-rank plus
  diagonal form; Local methods where nearby data is used to
  make predictions in a given region of space; and fast matrix-vector
  multiplication (MVM) methods, which can be used with iterative
  methods to speed up the solution of linear systems.  We discuss
  these in turn, so as to give coverage to the wide variety methods
  that have been proposed. We use the Fully Independent Training
  Conditional (FITC) method as it is recommended over other
  inducing point methods in \citet{candela2007}, and the
  Improved Fast Gauss Transform (IFGT) of \citet{yang2004} as a
  representative of fast MVM methods.

\subsection{Subset of Data}
\label{sec:sod}
The simplest way of dealing with large amounts of data is simply to
ignore some or most of it. The `Subset of Data (SoD) approximation'
simply applies the full GP prediction method to a subset of size $m <
n$. Therefore the computational complexities of SoD result from
replacing $n$ with~$m$ in the expressions for the full method
(Table~\ref{t:approxgpr}). Despite the `obvious' nature of SoD, most
papers on approximate GP methods only compare to a GP applied to the full
dataset of size~$n$.

To complete the description of the SoD method we must also specify how
the subset is selected. We consider two of the possible alternatives:
1)~Selecting $m$ points randomly costs $O(m)$ if we need not look at
the other points. 2)~We select $m$ cluster centres from a Farthest
Point Clustering (FPC, \citealt{gonzales-85}) of the dataset; using
the algorithm proposed by Gonzales this has
computational complexity of $O(mn)$. In theory, FPC can be
sped up to $O(n \log m)$ using suitable data structures
\citep{feder-greene-88}, although in practice the original algorithm
can be faster for machine learning problems of moderate
dimensionality. FPC has a random aspect as the first point can be
chosen randomly.
Our SoD implementation is based on \texttt{gp.m} in the
\matlab\ \texttt{gpml} toolbox:\\
\url{http://www.gaussianprocess.org/gpml/code/matlab/doc/}.

Rather than selecting the subset randomly, it is also possible to make
a more informed choice. For example
\citet{lawrence-seeger-herbrich-03} came up with a fast selection
scheme (the ``informative vector machine'') that takes only $O(m^2
n)$.  \citet{keerthi-chu-06} also proposed a matching pursuit approach
which has similar asymptotic complexity, although
the associated constant is larger.

\subsection{Inducing point methods: FITC}
\label{sec:fitc}
A number of GP approximation algorithms use alternative kernel matrices based
on
\emph{inducing points}, $\bu$, in the $D$-dimensional input space~\citep{candela2005}.
Here we restrict the $m$ inducing points to be a subset of the training
inputs. The Subset of Regressors (SoR) kernel function is given by
$k_{SoR}(\bx_i,\bx_j) = \bk(\bx_i,\bu) K^{-1}_{\bu \bu} \bk(\bu, \bx_j)$,
and the Fully Independent Training Conditional (FITC) method uses
\[ k_{FITC}(\bx_i,\bx_j) = k_{SoR}(\bx_i,\bx_j) + \delta_{ij} [k(\bx_i,\bx_j) -
k_{SoR}(\bx_i,\bx_j)]. \]
FITC approximates the matrix $K$ as a rank-$m$ plus diagonal matrix.
An attractive property of FITC, not shared by all approximations, is
that it corresponds to exact inference for a GP with the given
$k_{FITC}$ kernel \citep{candela2007}. Other inducing point
approximations (e.g.\ SoR, deterministic training conditionals) have
similar complexity but \citet{candela2007} recommend FITC over them.
Since then there have been further developments
\citep{titsias-09,lazarogredilla2010},
which would also be interesting to compare.

To make predictions with FITC, and to evaluate its
marginal likelihood, simply substitute $k_{FITC}$ for the
original kernel in Equations~\ref{eq:meanpred}--\ref{eq:logmarglike}.
This substitution gives a mean predictor of the form
$\overline{f}_* = \sum_{i=1}^m \beta_i k(\bx_*, \bx_i)$, where
$i = 1, \ldots, m$ indexes the selected subset of training points,
and the $\beta$s are obtained by solving a linear system.
\citet[pp 60-62]{snelson2007} showed that in
the limit of zero noise FITC reduces to SoD\@.

We again choose a set of inducing points of size $m$ from the
training inputs either randomly or using FPC, and use the FITC
implementation from the \texttt{gpml} toolbox.

It is possible to ``mix and match'' the SoD and FITC  methods,
adapting the hyperparmeters to optimize the SoD approximation to
the marginal likelihood, then using the FITC algorithm to make
predictions using the same data subset and the SoD-trained hyperparameters.
We refer to this procedure as the Hybrid method\footnote{We thank one of
the anonymous reviewers for suggesting this method.}. We expect that saving time
on the hyperparameter learning phase, $O(m^3)$ instead of $O(m^2n)$, will come
at the cost of reducing the predictive performance of FITC for a given~$m$.

\subsection{Local GPR}
\label{sec:localgpr}
The basic idea here is of divide-and-conquer, although without any
guarantees of correctness.
We divide the $n$ training points into $k=\lceil \frac{n}{m} \rceil$ clusters
each of size $m$, and run
GPR in each cluster, ignoring the training data outside of the given
cluster. At test time we assign a test input $\bx_*$ to the closest
cluster. This method has been discussed by
\citet{snelson-ghahramani-07}. The hard cluster boundaries can lead to
ugly discontinuities in the predictions, which are unacceptable if a
smooth surface is required, for example in some physical simulations.

One important issue is how the clustering is done. We found that
FPC tended to produce clusters of very unequal size, which
limited the speedups obtained by Local GPR\@. Thus we
devised a method we call Recursive Projection
Clustering (RPC), which works as follows. We start off with all the
data in one cluster $C$. Choose two data points at random from $C$,
draw a line through these points and calculate the orthogonal
projection of all points from $C$ onto the line. Split $C$ into
two equal-sized subsets $C_L$ and $C_R$ depending on whether
points are to the left or right of the median. Now repeat
recursively in each cluster until the cluster size is \cut{$\le m$.}
no larger than $m$.
In our implementation we make use of \matlab's \texttt{sort} function
to find the median value, taking time $O(n \log n)$ for $n$
datapoints, although it is possible to
reduce median finding to $O(n)$ \citep{blum-etal-73}. Thus overall
the complexity of RPC is $O(ns \log n)$, where $s=\lceil \log_2(n/m)
\rceil$. A test point $\bx_*$ is assigned to the appropriate
cluster by descending the tree of splits constructed by RPC\@.

Another issue concerns hyperparameter learning. $L$ is approximated
by the sum of terms like Eq.~\ref{eq:logmarglike}
over all clusters.  Hyperparameters can either be
tied across all clusters (``joint'' training), or unique to each
cluster (``separate'' training). Joint training is likely to be useful
for small $m$.
We implemented Local GPR using the \texttt{gpml} toolbox with small modifications to sum gradients for joint training.

\subsection{Iterative methods and IFGT matrix-vector multiplies}
\label{sec:itmvm}
The Conjugate Gradients (CG) method (e.g.,~\citealt{golub1996}) can be
used at training time to solve the linear system $(K + \sigma^2 I)
\balpha = \by$. Indeed, all GPR computations can be based on iterative
methods~\citep{gibbs1997}. CG~and several other iterative
methods (e.g., \citealt{li2007,liberty2007}) for solving linear
systems require the ability to
multiply a matrix of kernel values with an arbitrary vector.

Standard dense matrix-vector multiplication (MVM) costs $O(n^2)$.
It has been argued (e.g., \citealt{gibbs1997,li2007}) that
iterative methods alone provide a cost saving if terminated after
$k\ll n$ matrix-vector multiplies.
Papers often don't state how CG was terminated \citeg{shen2005,freitas2005},
although some are explicit about using a small fixed number of iterations based
on preliminary runs \citeg{gray2004a}. Ad-hoc termination rules, or those using
the `relative residual' \citep{golub1996} (see Section~\ref{sec:itresults}) do
not necessarily give the best trade-off between time and test-error. In
Section~\ref{sec:itresults} we examine the progression of test error throughout
training, to see what error/time trade-offs might be achieved by different
termination rules.

Iterative methods are not used routinely for dense linear system
solving, they are usually only recommended when the cost of MVMs is
reduced by exploiting sparsity or other matrix structure. Whether
iterative methods can provide a speedup for GPR or not, fast MVM
methods will certainly be required to scale to huge datasets. Firstly,
while other methods can be made linear in the size of the dataset size
($O(m^2n)$, see Table~\ref{t:approxgpr}), a standard
MVM costs $O(n^2)$. Most
importantly, explicitly constructing the $K$ matrix uses $O(n^2)$
memory, which sets a hard ceiling on dataset size. Storing the kernel
elements on disk, or reproducing the kernel computations on the fly,
is prohibitively expensive. Fast MVM methods potentially reduce the
storage required, as well as the computation time of the standard dense
implementation.

We have previously demonstrated some negative results
concerning speeding up MVMs \citep{murray-09}: 1)~if the kernel matrix were
approximately sparse (i.e.\ many entries near zero) it would be
possible to speed up MVMs using sparse matrix techniques, but in the
hyperparameter regimes identified in practice this does not usually
occur; 2)~the piecewise constant approximations used by simple kd-tree
approximations to GPR \citep{shen2005,gray2004a,freitas2005} cannot
safely provide meaningful speedups.

The Improved Fast Gauss Transform (IFGT) is a MVM method that can be
applied when using a squared-exponential kernel. The IFGT is based on
a truncated multivariate Taylor series around a number of cluster
centres. It has been applied to kernel machines in a number of
publications, e.g.\ \citep{yang2004,morariu-etal-09}.
Our experiments use the IFGT implementation from the Figtree C++
package with \matlab\ wrappers available from
\url{http://www.umiacs.umd.edu/~morariu/figtree/}. This software provides
automatic choices for a number of parameters  within IFGT\@.
The time complexity of IFGT depends on a number of factors as
described in \citep{morariu-etal-09}, and we focus below on empirical
results.

There are open problems with making iterative methods and fast MVMs
for GPR work routinely. Firstly, unlike standard dense linear algebra
routines, the number of operations depends on the hyperparameter
settings. Sometimes the programs can take a very long time, or even crash
due to numerical problems. Methods to diagnose and handle these
situations automatically are required. Secondly, iterative methods for
GPR are usually only applied to mean prediction,
Eq.~\ref{eq:meanpred}; finding variances $\mathbbm{V}[f_*]$ would
require solving a new linear system for each $\bk(\bx_*)$. In
principle, an iterative method could approximately factorize $(K +
\sigma^2 I)$ for variance prediction. To our knowledge, no one has
demonstrated the use of such a method for GPR with good scaling in
practice.

\subsection{Comparing the Approximation Methods}
Above we have reviewed the SoD, FITC, Hybrid, Local and Iterative MVM
methods for speeding up GP regression for large $n$. The space and
time complexities for the SoD, FITC, and Local methods are given
in Table~\ref{t:approxgpr}; as explained above there are
open problems with making iterative methods and fast MVMs work
routinely for GPR, see also Secs.\ \ref{sec:itresults} and
\ref{sec:ifgt_res}.

Comparing FITC to SoD, we note that the mean predictor contains the
same basis functions as the SoD predictor, but that the coefficients
are (in general) different as FITC has ``absorbed'' the effect of the
remaining $n-m$ datapoints. Hence for fixed $m$ we might expect FITC
to obtain better results. Comparing Local to SoD, we might expect that
using training points lying nearer to the test point would help, so
that for fixed $m$ Local would beat SoD\@. However, both FITC and Local
have $O(m^2 n)$ training times (although the associated constants may
differ), compared to $O(m^3)$ for SoD\@. So if equal training time was
allowed, a larger $m$ could be afforded for SoD than the others. This
is the key to the comparisons in Sec.\ \ref{sec:compsodfitclocal}
below. The Hybrid method has the same hyperparameter learning
time as SoD by definition, but the training phase will take longer than SoD
with the same $m$, because of the need for a final $O(m^2n)$ phase of
FITC training, as compared to the $O(m^3)$ for SoD\@. However, as per
the argument above, we would expect the FITC predictions to be superior to
the SoD ones, even if the hyperparameters have not been optimized
explicitly for FITC prediction; this is explored experimentally in Sec.\
\ref{sec:compsodfitclocal}.

At test time Table~\ref{t:approxgpr} shows that the SoD, FITC, Hybrid
and Local approximations are $O(m)$ for mean prediction, and $O(m^2)$
for predictive variances. This means that the method which has
obtained the best ``$m$-size'' predictor will win on test-time
performance.

\section{A Basis for Comparing Approximations \label{sec:compare}}
For fixed hyperparameters, comparing an approximate method to the full
GPR is relatively straightforward: we can evaluate the predictive
error made by the approximate method, and compare that against the
``gold standard'' of full GPR\@. The `best' method could be the
approximation with best predictions for a given computational cost, or
alternatively the smallest computational cost for a given predictive
performance.  However, there are still some options e.g.\ different
performance criteria to choose from (mean squared error, mean
predictive log likelihood). Also there are different possible relevant
computational costs (hyperparameter learning, training, testing) and
definitions of cost itself (CPU time, `flops' or other operation
counts). It should also be borne in mind that any error measure
compresses the predictive mean and variance functions into a single
number; for low-dimensional problems it is possible to visualize these
functions, see e.g.\ Fig.\ 9.4 in \citet{candela2007}, in order to
help understand the differences between approximations.

It is rare that the appropriate hyperparameters are known for
a given problem, unless it is a synthetic problem drawn from a GP\@.
For real-world data we are faced with two alternatives: (i)~compare
approximate methods using the same set of hyperparameters as obtained
by full GPR, or (ii)~allow the approximate methods freedom
to determine their own hyperparameters, e.g.\ by using approximate
marginal likelihoods consistent with the approximations. Below we follow
the second approach as it is more realistic, although it does
complicate comparisons by changing both the approximation method and
the hyperparameters.

In terms of computational cost we use the CPU time in
seconds, based on \matlab\ implementations of the algorithms
(except for the IFGT where the Figtree C++ code is used with
\matlab\ wrappers). The core GPR calculations are well suited to
efficient implementation in \matlab. Our SoD, FITC, Hybrid and Local GP
implementations are all derived from the standard \texttt{gpml}
toolbox of Rasmussen and Nickisch.

Before making empirical comparisons on particular datasets, we
identify aspects of regression problems, models and approximations
that affect the appropriateness of using a
particular method:

\textbf{The nature of the underlying problem:} We
usually standardize the inputs
to have zero mean and unit variance on each dimension. Then
clearly we would expect to require more datapoints to pin down
accurately a higher frequency (more ``wiggly'') function than
a lower frequency one.

For multivariate input spaces there will also be issues of
dimensionality, either wrt the intrinsic dimensionality of $\bx$ (for
example if the data lies on a manifold of lower dimensionality) or the
apparent dimensionality. Note that if there are irrelevant inputs
these can potentially be detected by a kernel equipped with
``Automatic Relevance Determination'' (ARD)
(\citealp{neal-96}; \citealp[p.~106]{rasmussen-williams-06}).

Another factor is the noise level on the
data. An eigenanalysis of the problem (see e.g.\
\citealt[~Sec.~2.6]{rasmussen-williams-06}) shows that it is more
difficult to discover low-amplitude components in the
underlying function if there is high noise.
It is relatively easy to get an upper bound on the noise
level by computing the variance of the $y$'s around a given
$\bx$ location (or an average of such calculations),
particularly if the lengthscale of variation of function is much larger
than inter-datapoint distances (i.e.\ high sampling density); this
provides a useful sanity check on the noise level returned
during hyperparameter optimization.

\textbf{The choice of kernel function:} Selecting an appropriate
family of kernel functions is an important part of modelling a
particular problem. For example, poor results can be obtained when
using an isotropic kernel on a problem where there are irrelevant
input dimensions, while an ARD parameterization would be a better
choice.  Some approximation methods (e.g., the IFGT) have only been
derived for particular kernel functions.  For simplicity of comparison
we consider only the SE-ARD kernel \cite[p.~106]{rasmussen-williams-06},
as that is the kernel most widely used in practice.

\textbf{The practical usability of a method:}
Finally, some more mundane issues contribute significantly to the
usability of a method, such as:
(a)~Is the method numerically robust? If there are problems it
  should be clear how to diagnose and deal with them.
(b)~Is it clear how to set tweak parameters e.g.\ termination criteria?
    Difficulties with these issues don't just make it difficult to make
 fair comparisons, but reflect real difficulties with using the methods.
(c)~Does the method work efficiently for a wide range of hyperparameter
    settings? If not, hyperparameter searching must be performed much more
    carefully and one has to ask if the method will work well on good
    hyperparameter settings.

\section{Experiments \label{sec:expts}}
{\bf Datasets}: We use four datasets for comparison. The first two are synthetic
datasets, \textsc{synth2} and \textsc{synth8}, with $D\te2$ and
$D\te8$ input dimensions. The inputs were drawn from a $N(0,I)$
Gaussian, and the function was drawn from a GP with zero mean and
isotropic SE kernel with unit lengthscale. There are 30,543 training
points and 30,544 test points in each dataset.\footnote{We thank
Carl Rasmussen for providing these datasets.}
The noise variance is
$10^{-6}$ for \textsc{synth2}, and $10^{-3}$ for \textsc{synth8}.
The \textsc{chem} dataset is derived from physical simulations
relating to electron energies in molecules\footnote{We thank
Prof.\ Lionel Raff of Oklahoma State University and colleagues
for permission to distribute this data.} \citep{malshe-etal-05}.
The input dimensionality is 15, and the data is split into
31,535 training cases and 31,536 test cases. Additional results
on this dataset have been reported by \citet{manzhos2008}.
The \textsc{sarcos} dataset concerns the inverse kinematics of a robot
arm, and is used e.g.\ in \citealt*[Sec.~2.5]{rasmussen-williams-06}.
It has 21 input dimensions, 44,484 training cases and 4,449 test cases
(the same split as in \citealt{rasmussen-williams-06}).
The \textsc{sarcos} dataset is already publicly available from
\url{http://www.gaussianprocess.org}. All four datasets are included
in the code and data tarfile associated with this paper.

{\bf Error measures}: We measured the accuracy of the methods' predictions
on the test sets using the Standardized Mean Squared Error (SMSE), and
Mean Standardized Log Loss (MSLL), as defined
in \cite[Sec.~2.5]{rasmussen-williams-06}. The SMSE is the
mean squared error normalized by the MSE of the dumb predictor
that always predicts the mean of the training set. The MSLL is
obtained by
averaging $-\log p(y_*|{\cal D}, \bx_*)$ over the test
set  and subtracting the same score for a trivial model
which always predicts the mean and variance of the training set.
Notice that MSLL involves the predictive variances while SMSE does
not.

Each experiment was carried out on a 3.47\,GHz core with at least 10\,GB
available memory, except for Section~\ref{sec:itresults} which used 3\,GHz cores
with 12\,GB memory.
Approximate log marginal likelihoods were optimized
wrt $\theta$ using Carl Rasmussen's \texttt{minimize.m}
routine from the \texttt{gpml} toolbox, using a maximum
of 100 iterations.
The code and data used to run the experiments is
available from
\url{http://homepages.inf.ed.ac.uk/ckiw/code/gpr_approx.html} .

In Section~\ref{sec:itresults} we provide results investigating
the efficacy of iterative methods for GPR\@. In Section
\ref{sec:ifgt_res} we investigate the utility of IFGT to speed up
MVMs. Section~\ref{sec:compsodfitclocal} compares the SoD, FITC
and Local approximations on the four datasets, and Section
\ref{sec:compare_gen} compares predictions made with the learned
hyperparameters and the generative hyperparameters on the
synthetic datasets.

\subsection{Results for iterative methods \label{sec:itresults}}
Most attempts to use iterative methods for Gaussian processes have
used conjugate gradient (CG) methods
\citep{gibbs1997,gray2004a,shen2005,freitas2005}. However,
\citet{li2007} introduced a method, which they called Domain
Decomposition (DD), that over 50 iterations appeared to converge
faster than CG\@. We have compared CG and DD for training a GP
mean predictor based on 16,384 points from the \textsc{sarcos} data, with the
same fixed hyperparameters used by \citet{rasmussen-williams-06}.

\begin{figure}
\begin{minipage}{0.25\linewidth}
    \includegraphics[width=1.01\linewidth]{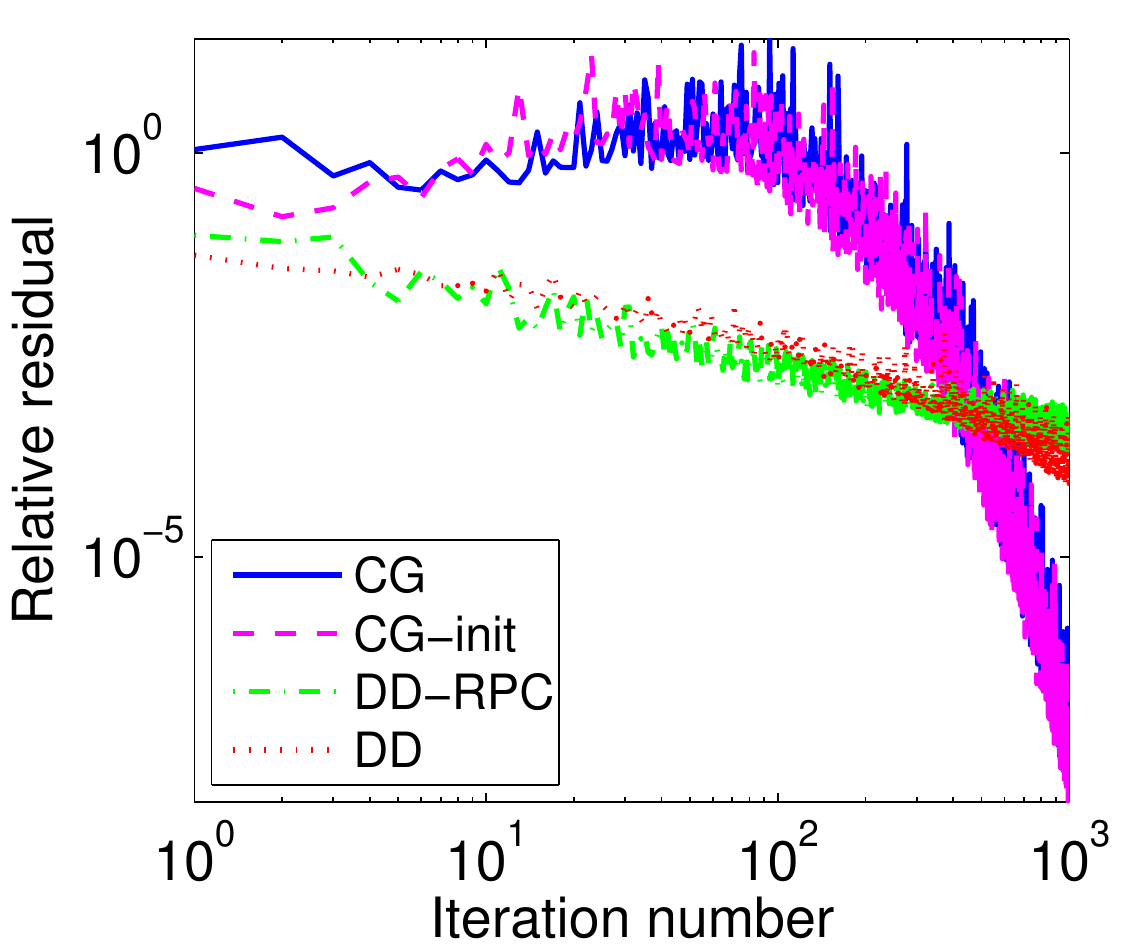}\\[1mm]
    \centerline{a) \textsc{sarcos}}
\end{minipage}%
\begin{minipage}{0.25\linewidth}
    \includegraphics[width=1.01\linewidth]{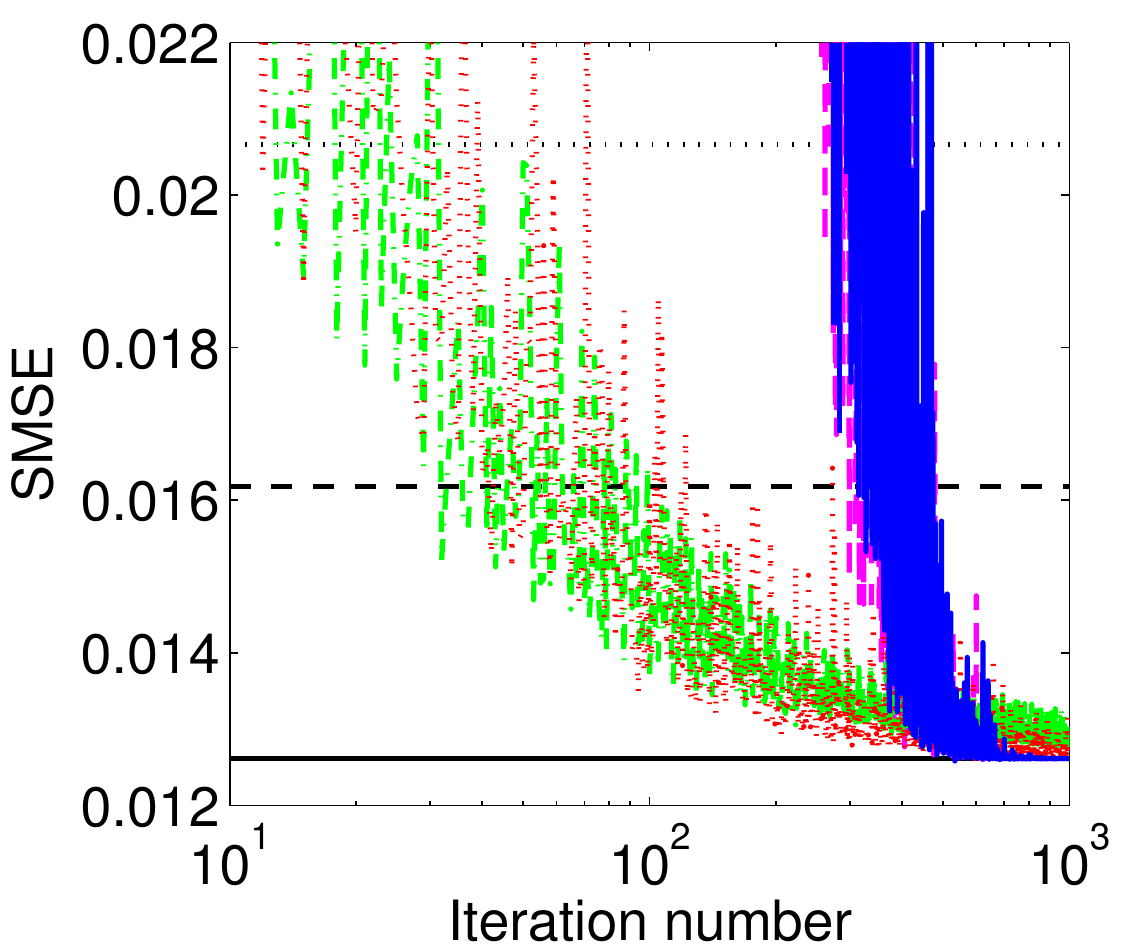}\\[1mm]
    \centerline{b) \textsc{sarcos}}
\end{minipage}%
\begin{minipage}{0.25\linewidth}
    \includegraphics[width=1.01\linewidth]{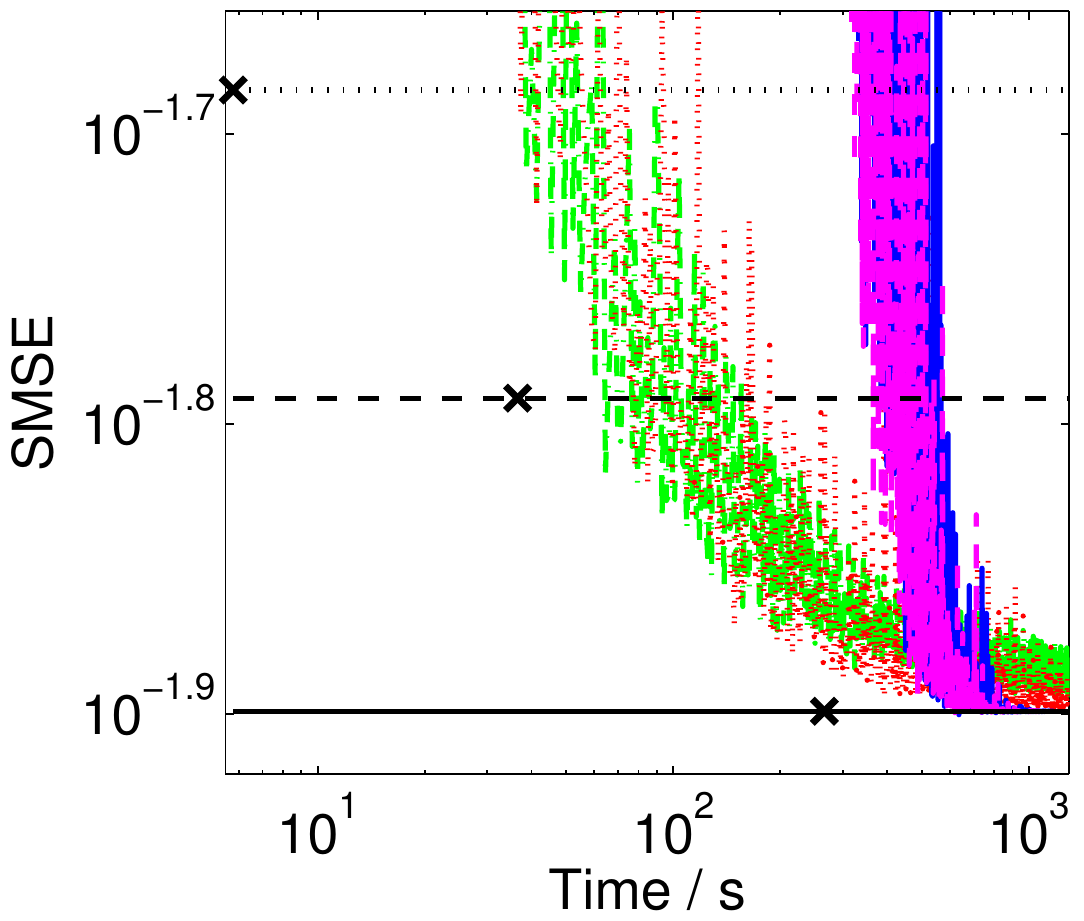}\\[1mm]
    \centerline{c) \textsc{sarcos}}
\end{minipage}%
\begin{minipage}{0.25\linewidth}
    \includegraphics[width=1.01\linewidth]{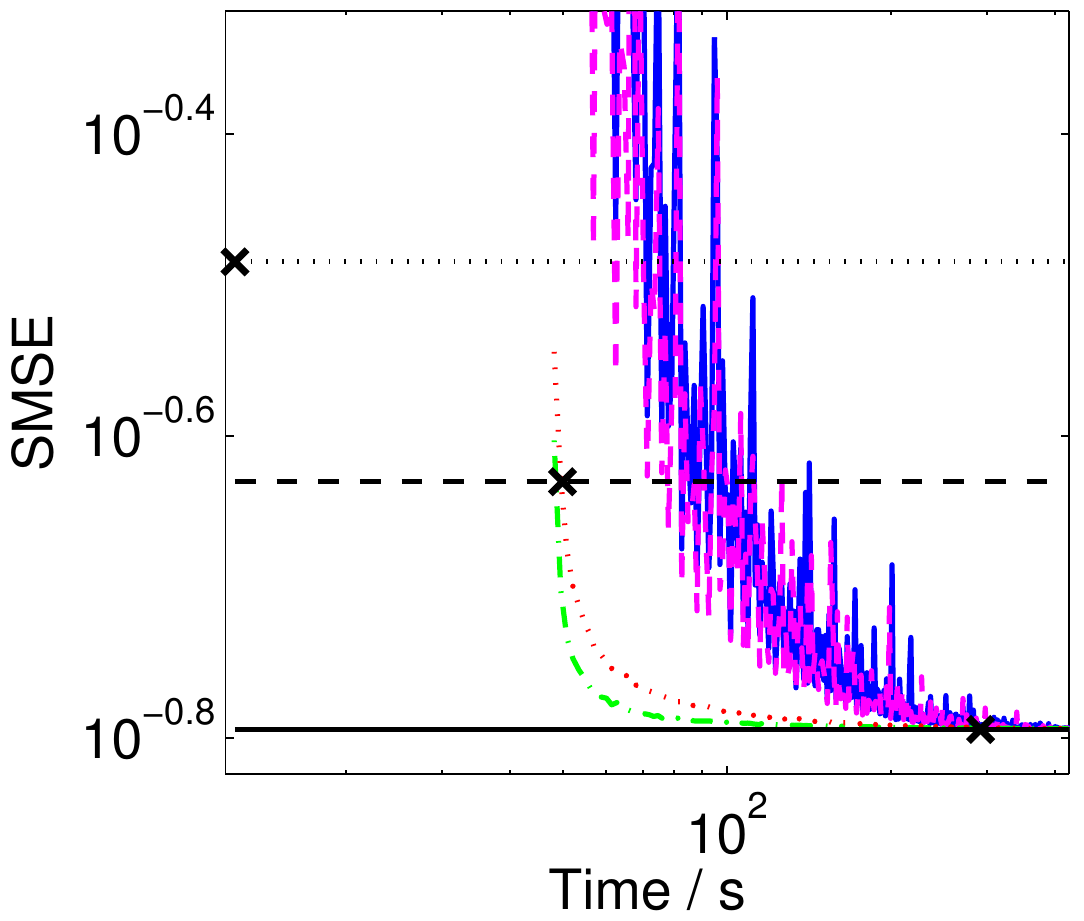}\\[1mm]
    \centerline{d) \textsc{synth8}}
\end{minipage}
\vspace*{-2.5mm}
\caption{Experiments with 16,384 training points. CG:
conjugate gradients; DD: `domain decomposition' with 16 randomly chosen
clusters; CG-init: CG initialized with one iteration of DD (CG's
starting point of zero isn't responsible for bad early behaviour);
DD-RPC: clusters were chosen with recursive projection clustering
(Section~\ref{sec:localgpr}). The horizontal lines give test
performance for SoD with 4,096, 8,192 and 16,384 training points.
Crosses on these lines also show the time taken.
}
\label{fig:iter}
\end{figure}

Figure~\ref{fig:iter}a) plots the `relative residual', $\|(K \tp
\sigma^2 I) \balpha_t - \by\| / \|\by\|$, the convergence diagnostic
used by \citet[Fig.~2]{li2007}, against iteration number for both
their method and~CG, where $\balpha_t$ is the approximation to
$\balpha$ obtained at iteration~$t$.
We reproduce the result that CG gives higher and fluctuating residuals
for early iterations. However, by running the simulation for longer,
and plotting on a log scale, we see that CG converges, according to
this measure, much faster at later iterations. Figure~\ref{fig:iter}a)
is not directly useful for choosing between the methods however, because we
do not know how many iterations are required for a competitive test-error.

Figure~\ref{fig:iter}b) instead plots test-set SMSE, and adds
reference lines for the SMSEs obtained by subsets with 4,096,
8,192 and 16,384 training points. We now see that 50 iterations are
insufficient for meaningful convergence on this problem.
Figure~\ref{fig:iter}c) plots the SMSE against computer time
taken on our machine\footnote{The time per iteration was measured on a
separate run that wasn't slowed down by storing the intermediate
results required for these plots.}. SoD performs better than the
iterative methods.

These results depend on the dataset and hyperparameters.
Figure~\ref{fig:iter}d) shows the test-set SMSE progression against
time for 16,384 points from \textsc{synth8} using the true
hyperparameters. Here CG takes a similar time to direct Cholesky
solving. However, there is now a part of the error-time plot where the
DD approach has better SMSEs at smaller times than either CG or
SoD\@.

The timing results are heavily implementation and architecture
dependent. For example, the results reported so far were run on a
single 3\,GHz core. On our machines, the iterative methods scale less
well when deployed on multiple CPU cores. Increasing the number of
cores to four (using \matlab, which uses Intel's~MKL), the time to
perform a $16384\!\times\!16384$ Cholesky decomposition decreased by a
factor of~3.1, whereas a matrix vector multiply improved by only a
factor of~1.7.

\subsection{Results for IFGT \label{sec:ifgt_res}}
\begin{figure}
\begin{minipage}{0.25\linewidth}
    \includegraphics[width=1.01\linewidth]{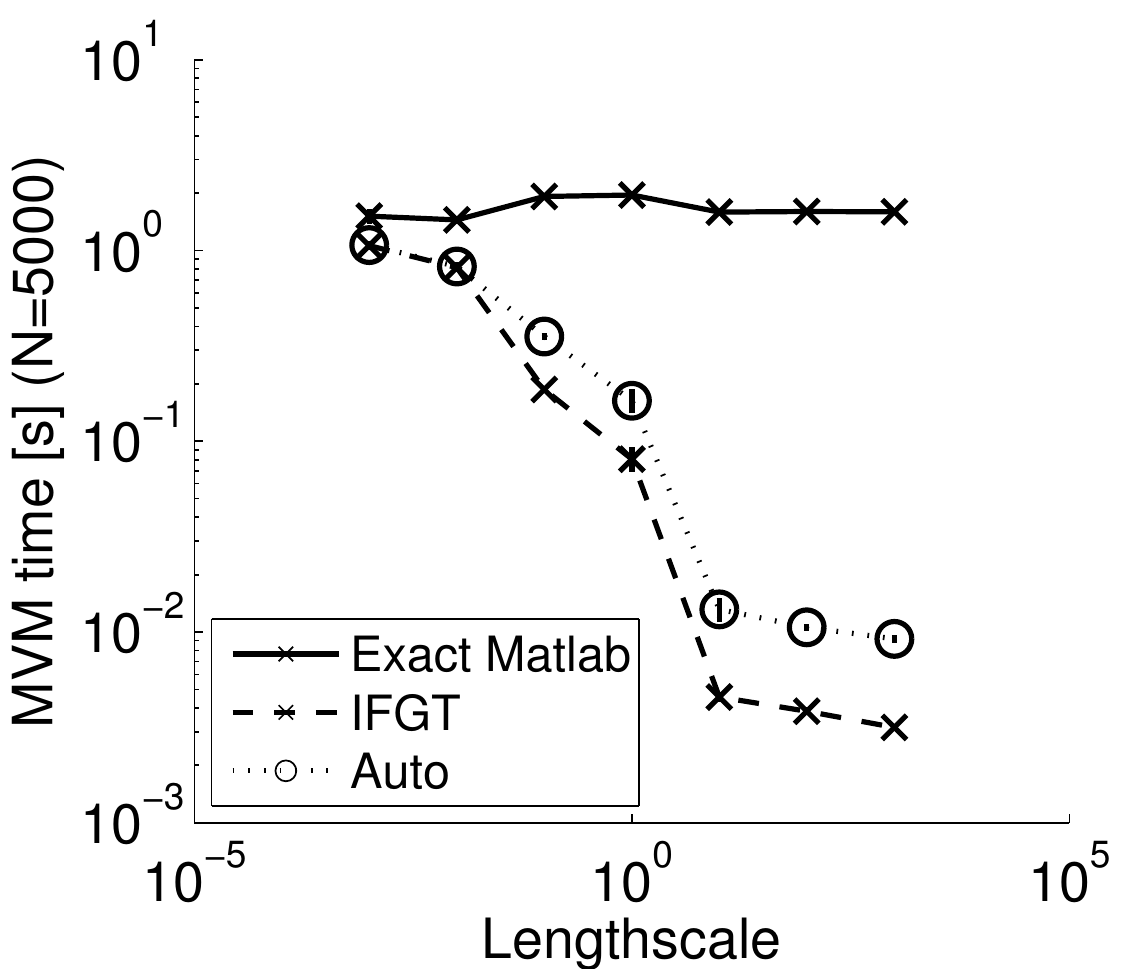}\\[1mm]
    \centerline{\textsc{synth2}}
\end{minipage}%
\begin{minipage}{0.25\linewidth}
    \includegraphics[width=1.01\linewidth]{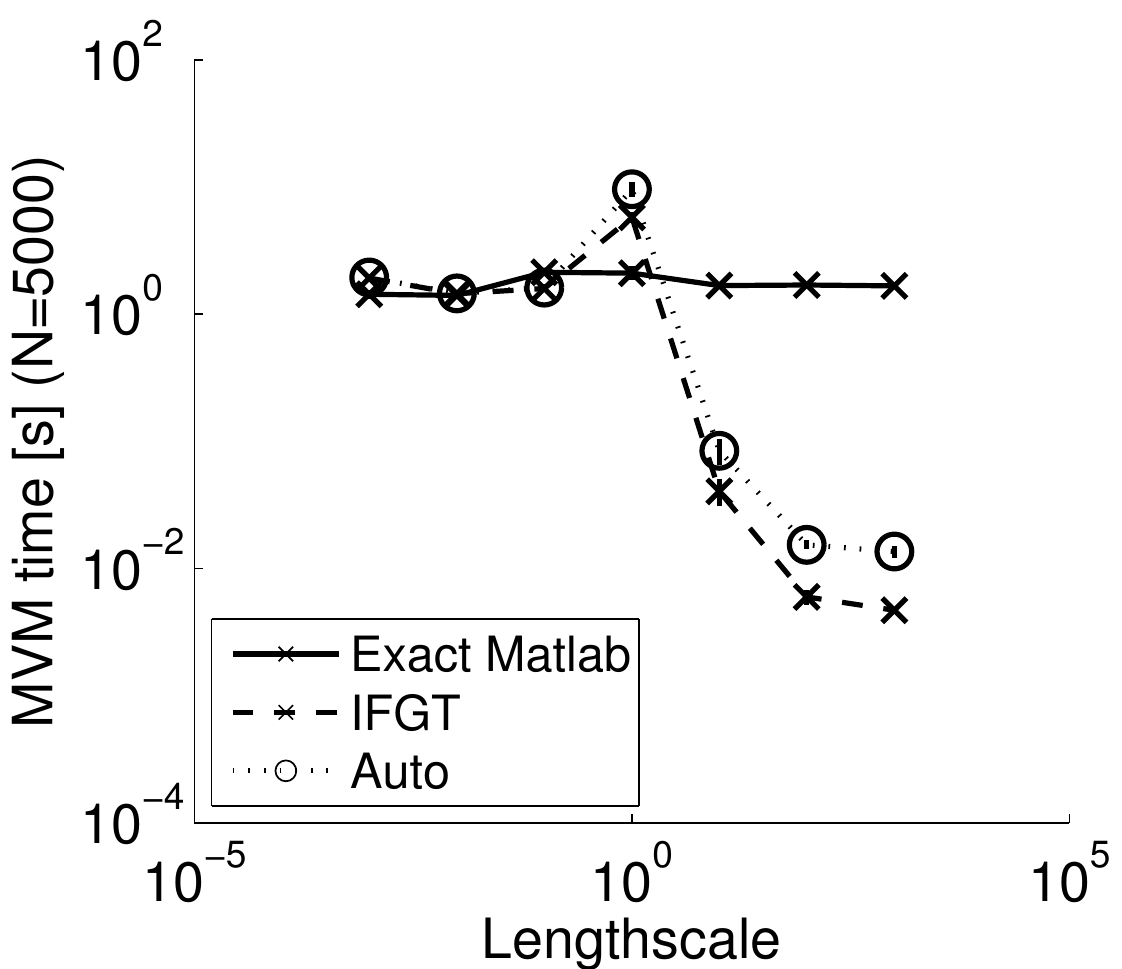}\\[1mm]
    \centerline{\textsc{synth8}}
\end{minipage}%
\begin{minipage}{0.25\linewidth}
    \includegraphics[width=1.01\linewidth]{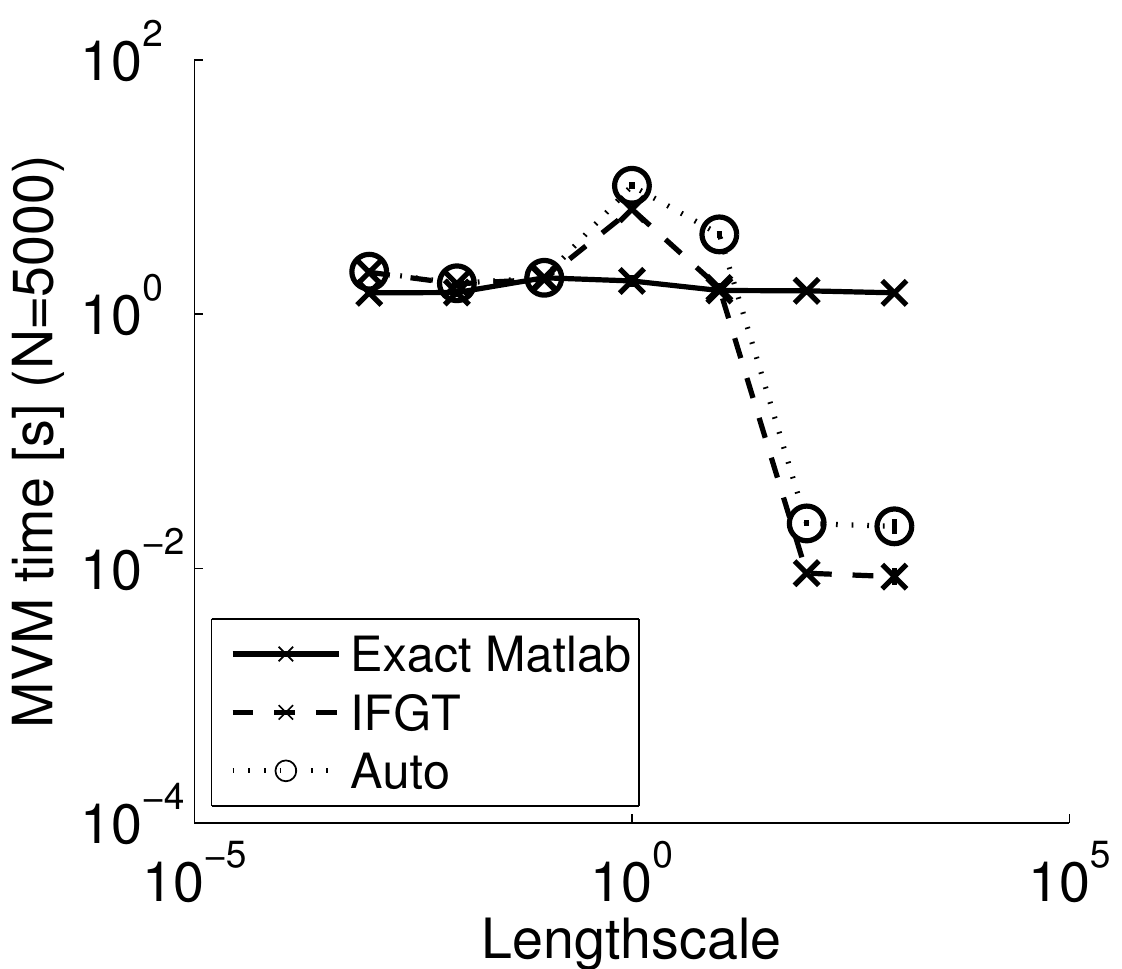}\\[1mm]
    \centerline{\textsc{chem}}
\end{minipage}%
\begin{minipage}{0.25\linewidth}
    \includegraphics[width=1.01\linewidth]{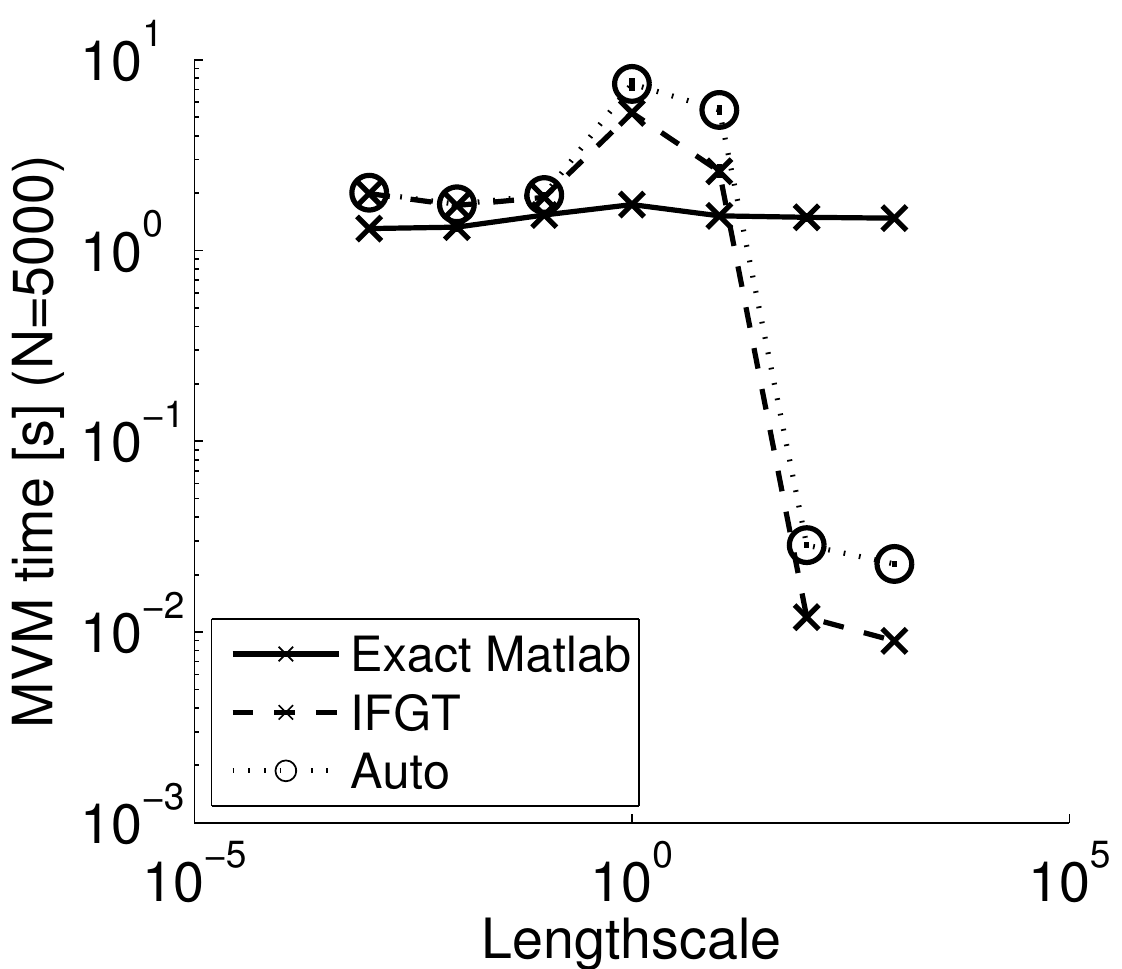}\\[1mm]
    \centerline{\textsc{sarcos}}
\end{minipage}
\vspace*{-2.5mm}
\caption{Plot of time vs lengthscale using IFGT for matrix-vector
multiplication (MVM) on the four
datasets. The Auto method was introduced in \citep{raykar2007} as a way
to speed up IFGT in some regimes.
\label{fig:mvm}}
\vspace*{-2mm}
\end{figure}

We focus here on whether the IFGT provides fast MVMs for the
datasets in our comparison.
We used the isotropic squared-exponential
kernel (which has one lengthscale parameter shared over all dimensions).
For each of the four datasets we randomly chose 5000 datapoints
to construct a kernel matrix, and a 5000-element random vector
(with elements sampled from $U[0,1]$). Figure~\ref{fig:mvm}
shows the MVM time as a function of lengthscale. For \textsc{synth2}
and \textsc{synth8} the known lengthscale is~1. For the two other problems, and
indeed many standardized regression problems,
lengthscales of~~$\approx\!1$ (the width of the input distribution)
are also appropriate. Figure~\ref{fig:mvm} shows that
useful MVM speedups over a direct implementation are only obtained
for \textsc{synth2}.
The result on \textsc{sarcos} is consistent with
\citet{raykar2007}'s result that
IFGT does not accelerate GPR on this dataset.

\subsection{Comparison of SoD, FITC, Hybrid and Local GPR \label{sec:compsodfitclocal}}
All of the experiments below used the squared exponential kernel with ARD
parameterization \hbox{\cite[p106]{rasmussen-williams-06}}. The test times given below
include computation of the predictive variances.

SoD was run with $m$ ascending in powers of 2 from
$32, 64 \ldots$ up to $4096$. FITC was run with $m$ ranging from $8$ to $512$ in powers of
two; this is smaller than for SoD as FITC is much more
memory intensive. Local was run with $m$ ranging from $16$ to $2048$ in
powers of two. For all experiments the selection of the subset/inducing
points/clusters has a random aspect, and
we performed five runs.

\begin{figure}
\begin{center}
\begin{tabular}{ccc}
\raisebox{4.5cm}{\textsc{synth2}} &
\hspace*{-0.42cm}\includegraphics[width=5.5cm]{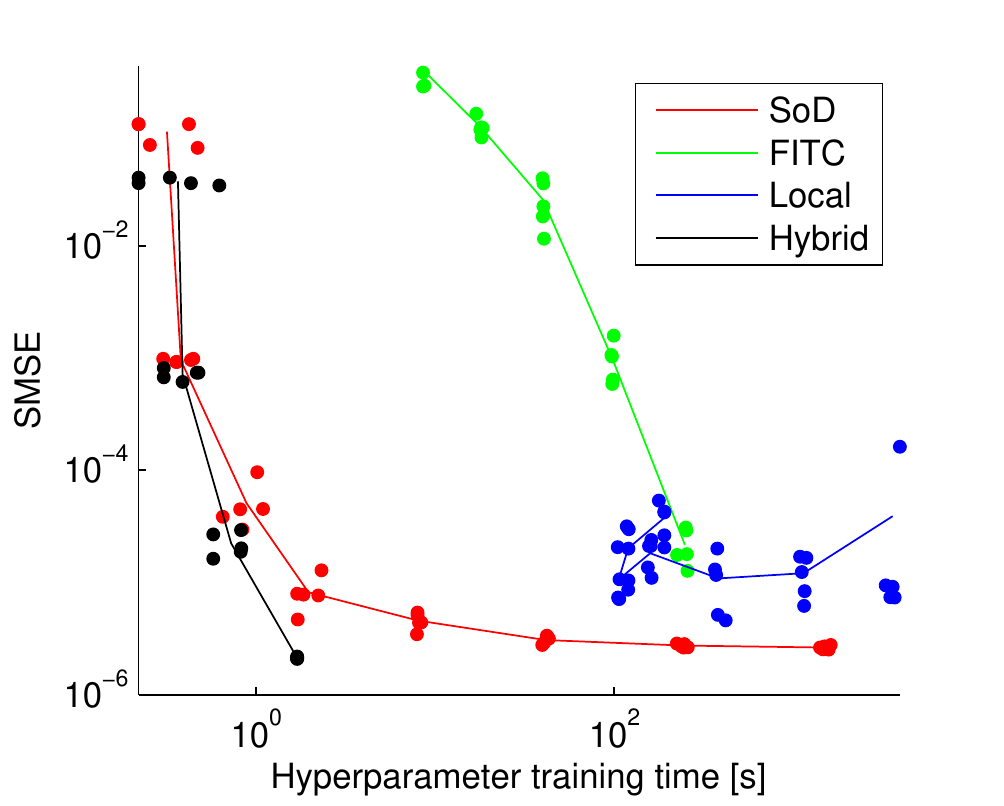}
\label{fig:SYNTH2HypSMSE} &
\hspace*{-0.41cm}\includegraphics[width=5.5cm]{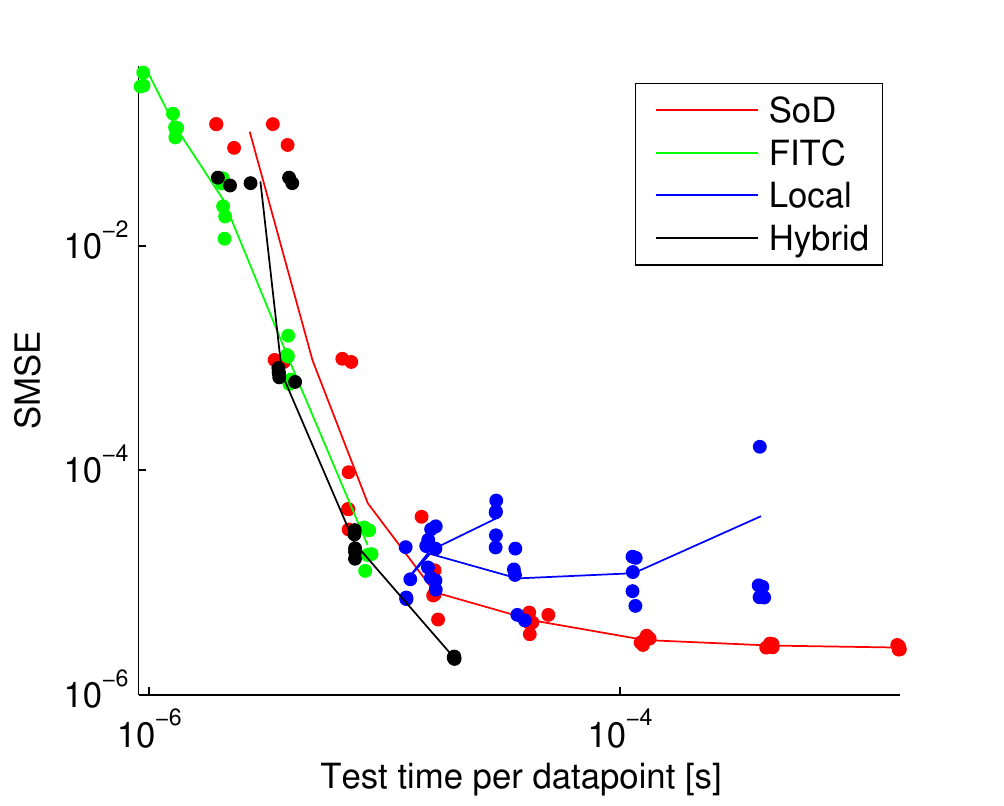}
\label{fig:SYNTH2TestSMSE} \\[-1mm]
\raisebox{4.5cm}{\textsc{synth8}} &
\hspace*{-0.65cm}\includegraphics[width=5.5cm]{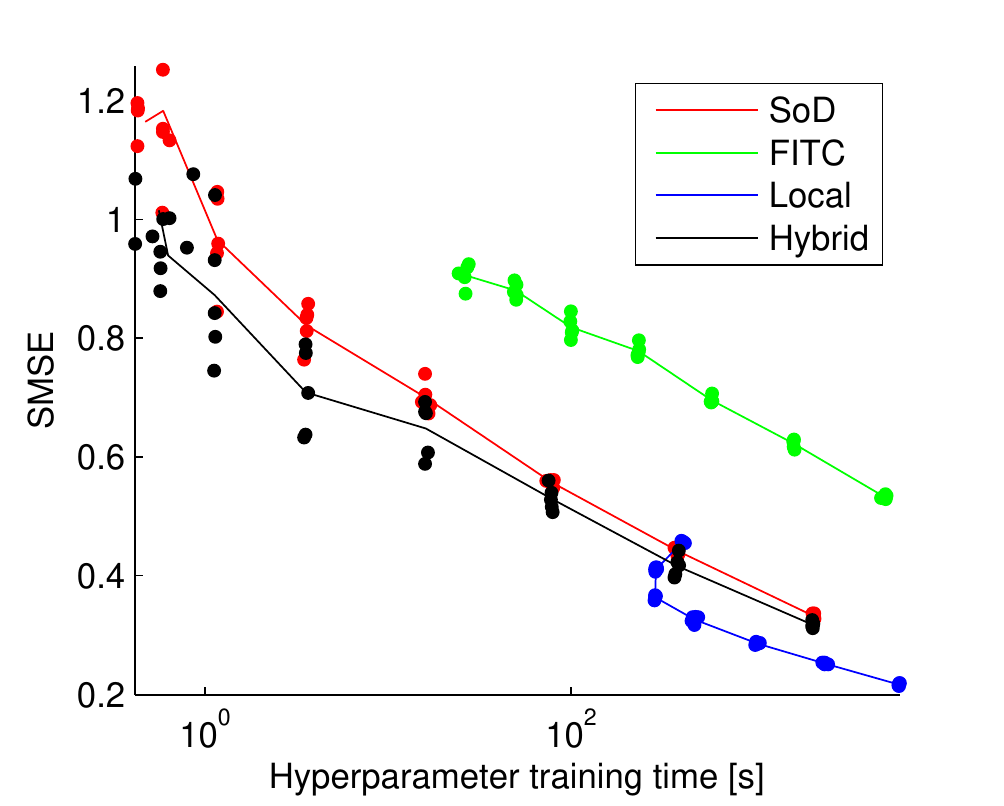}
\label{fig:SYNTH8HypSMSE} &
\hspace*{-0.65cm}\includegraphics[width=5.5cm]{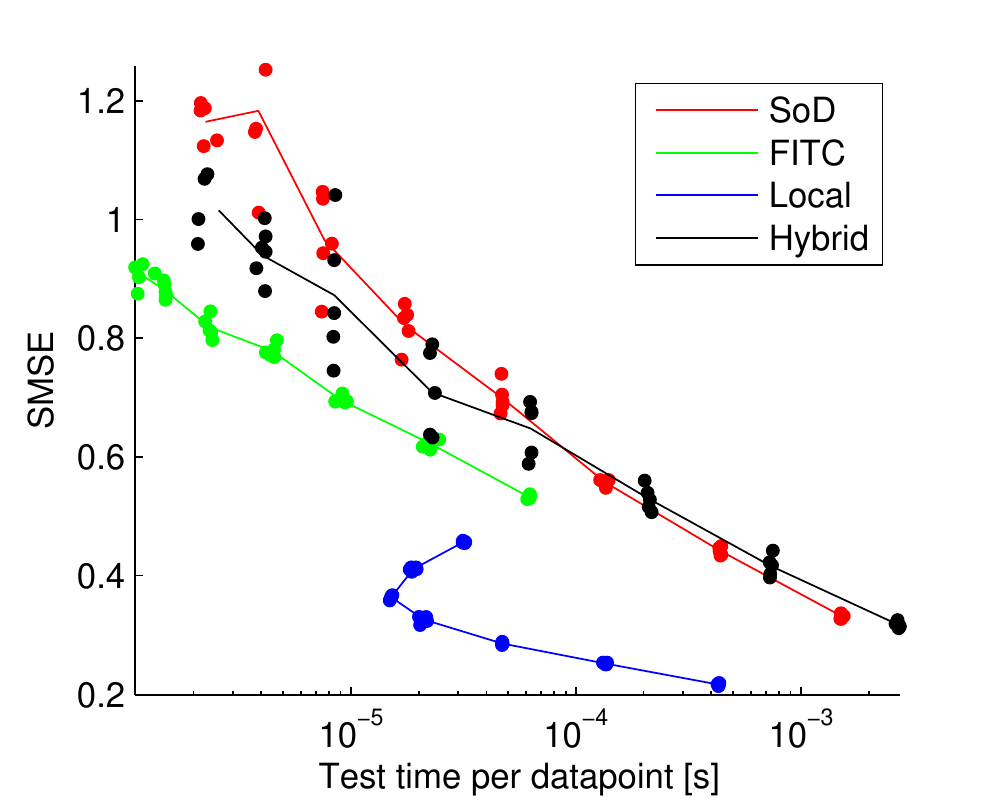}
\label{fig:SYNTH8TestSMSE} \\[-1mm]
\raisebox{4.5cm}{\textsc{chem}} &
\hspace*{-0.42cm}\includegraphics[width=5.5cm]{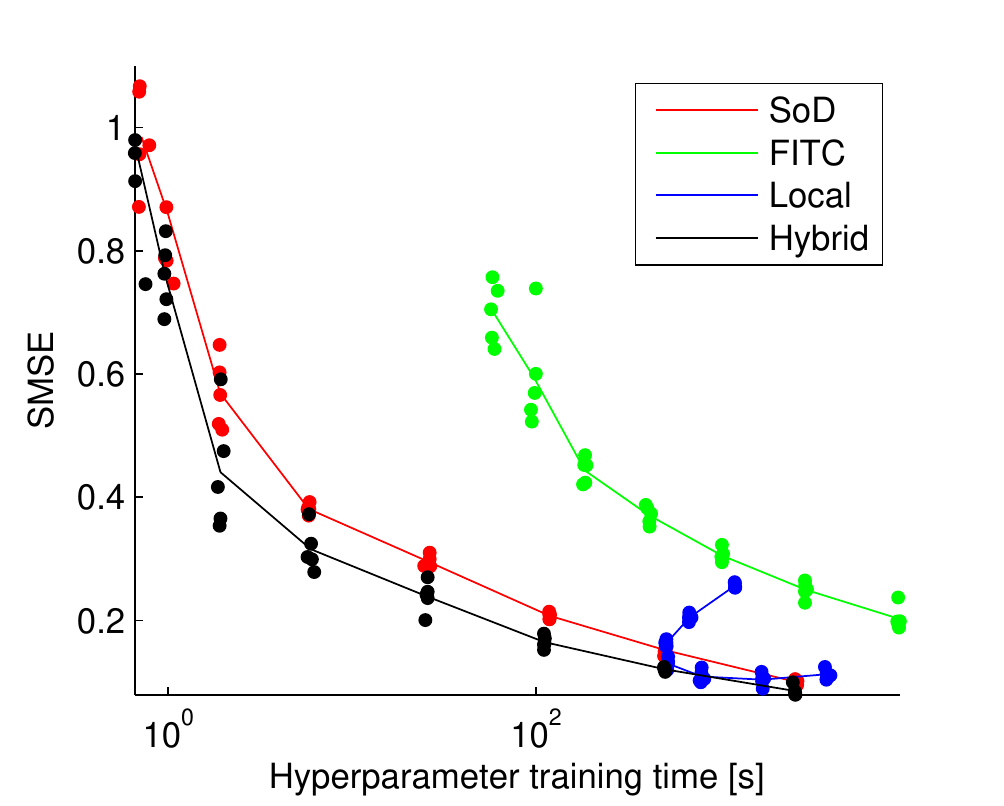}
\label{fig:CHEMHypSMSE} &
\hspace*{-0.41cm}\includegraphics[width=5.5cm]{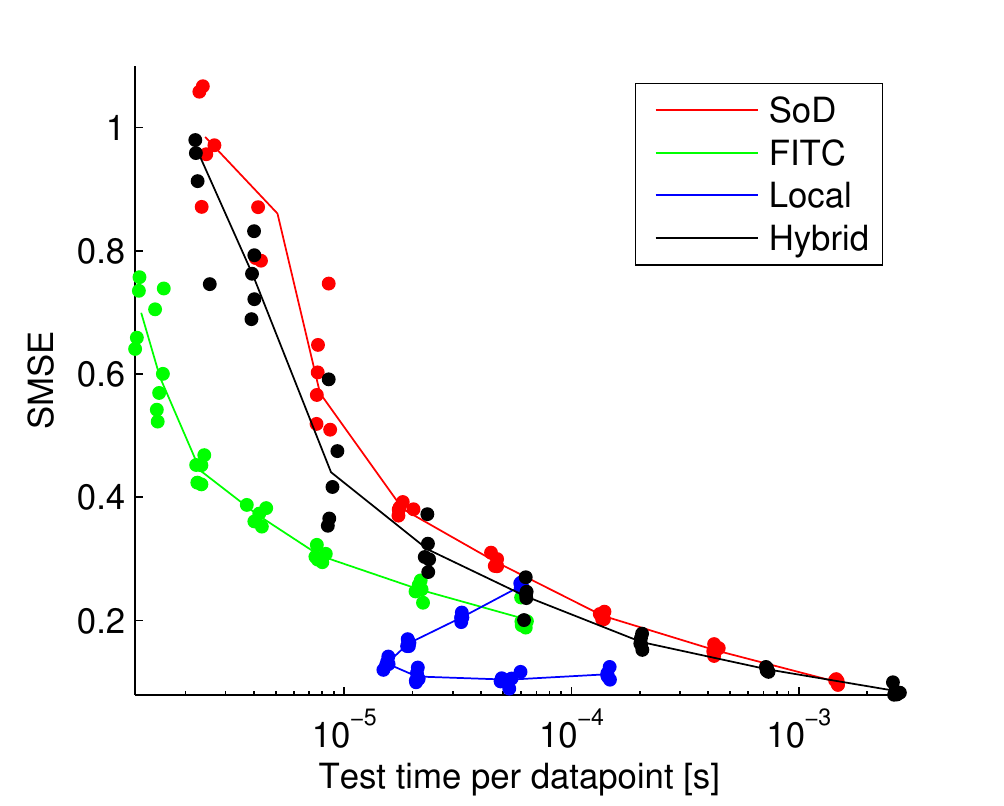}
\label{fig:CHEMTestSMSE} \\[-1mm]
\raisebox{4.5cm}{\textsc{sarcos}} &
\hspace*{-0.42cm}\includegraphics[width=5.5cm]{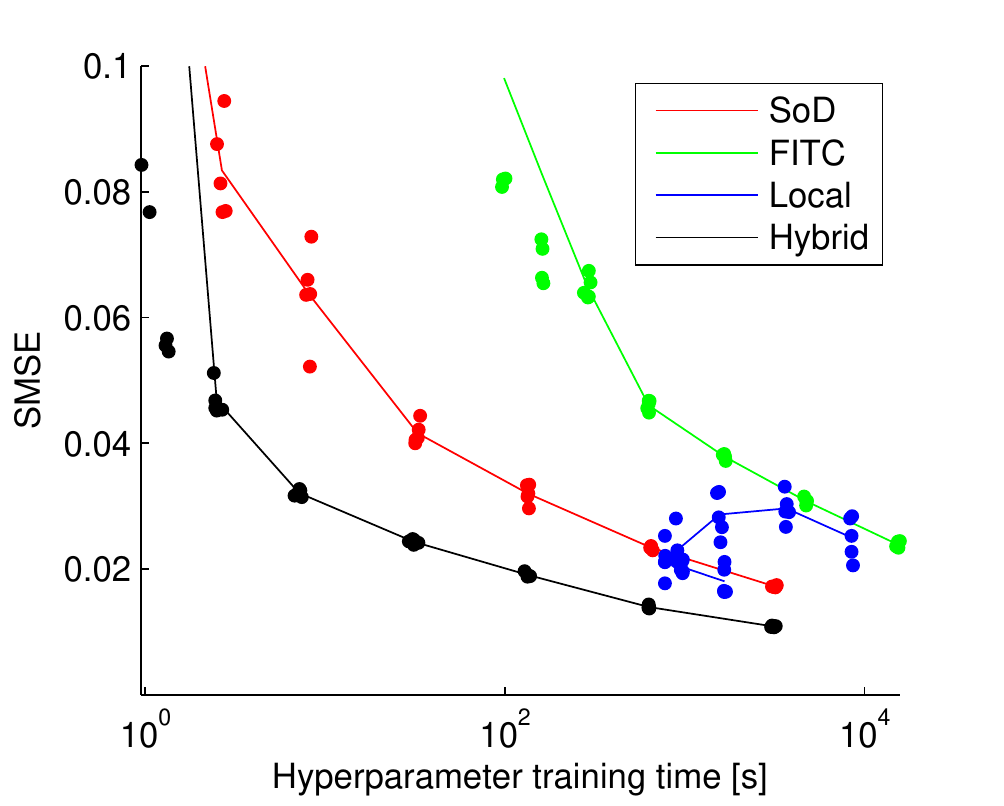}
\label{fig:SARCOSHypSMSE} &
\hspace*{-0.41cm}\includegraphics[width=5.5cm]{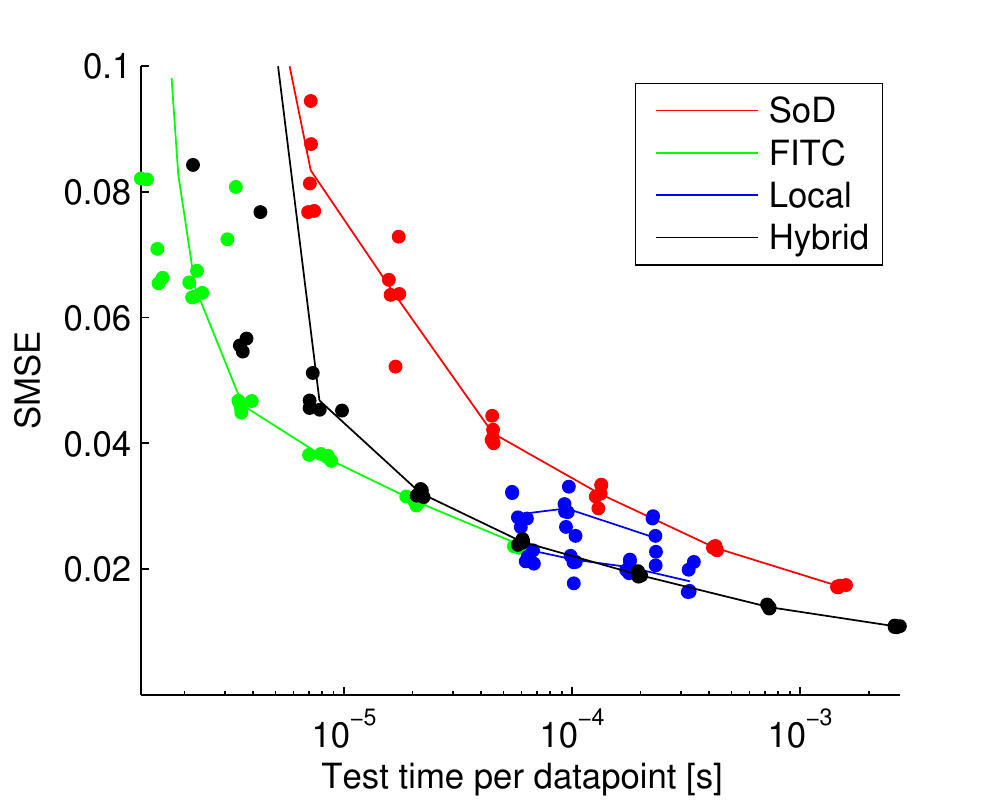}
\label{fig:SARCOSTestSMSE}
\end{tabular}
\end{center}
\caption{SMSE (log scale) as a function of time (log scale) for the
four datasets. Left: hyperparameter training
time. Right: test time per test point (including
variance computations, despite not being
needed to report SMSE)\@.
Points give the result for each run; lines connect the means of the 5 runs at each $m$.
\label{fig:timeres}}
\end{figure}

\begin{figure}
\begin{center}
\begin{tabular}{ccc}
\raisebox{4.5cm}{\textsc{synth2}} &
\hspace*{-0.42cm}\includegraphics[width=5.5cm]{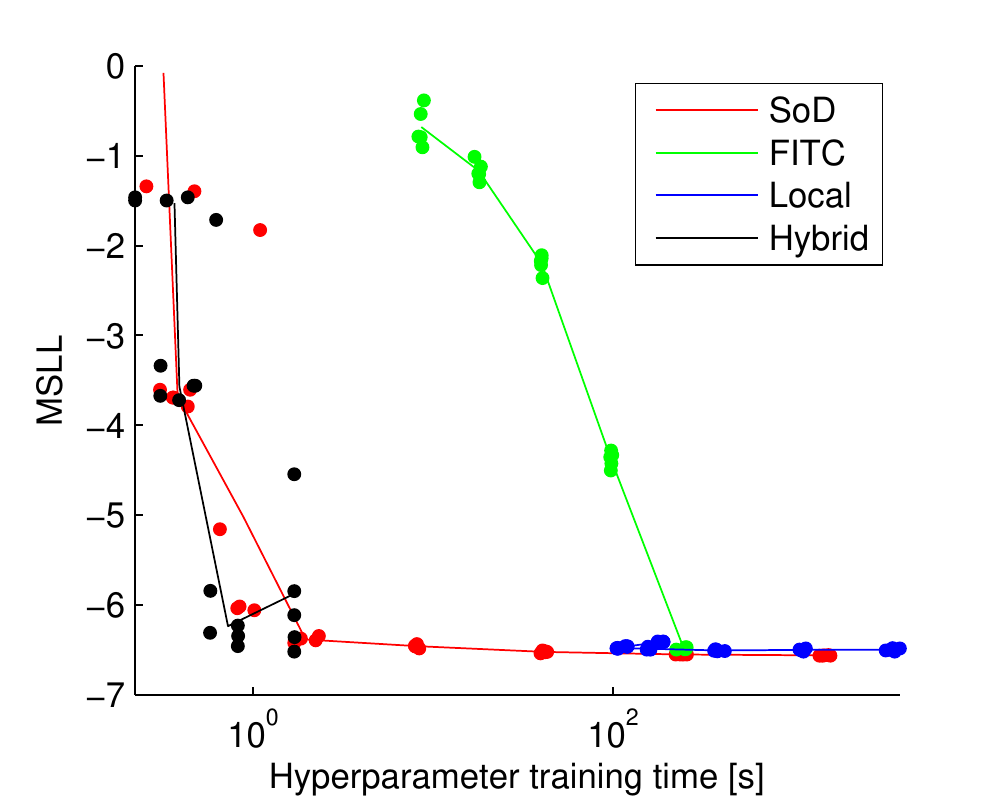}
\label{fig:SYNTH2HypMSLL} &
\hspace*{-0.41cm}\includegraphics[width=5.5cm]{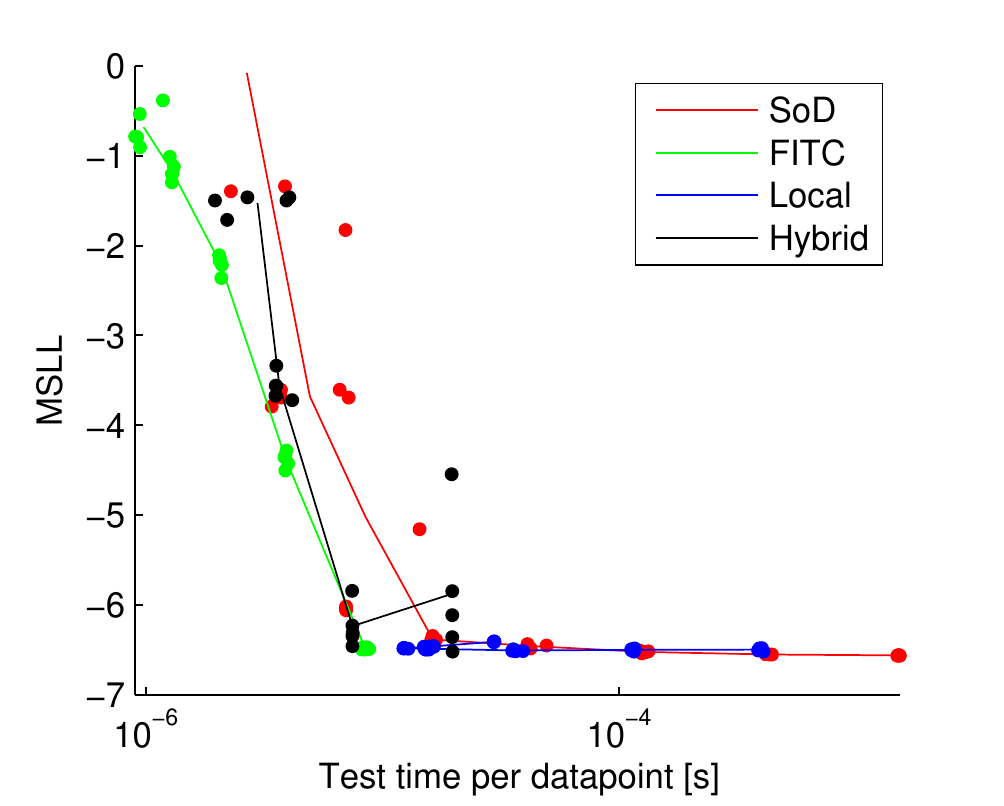}
\label{fig:SYNTH2TestMSLL} \\[-1mm]
\raisebox{4.5cm}{\textsc{synth8}} &
\hspace*{-0.65cm}\includegraphics[width=5.5cm]{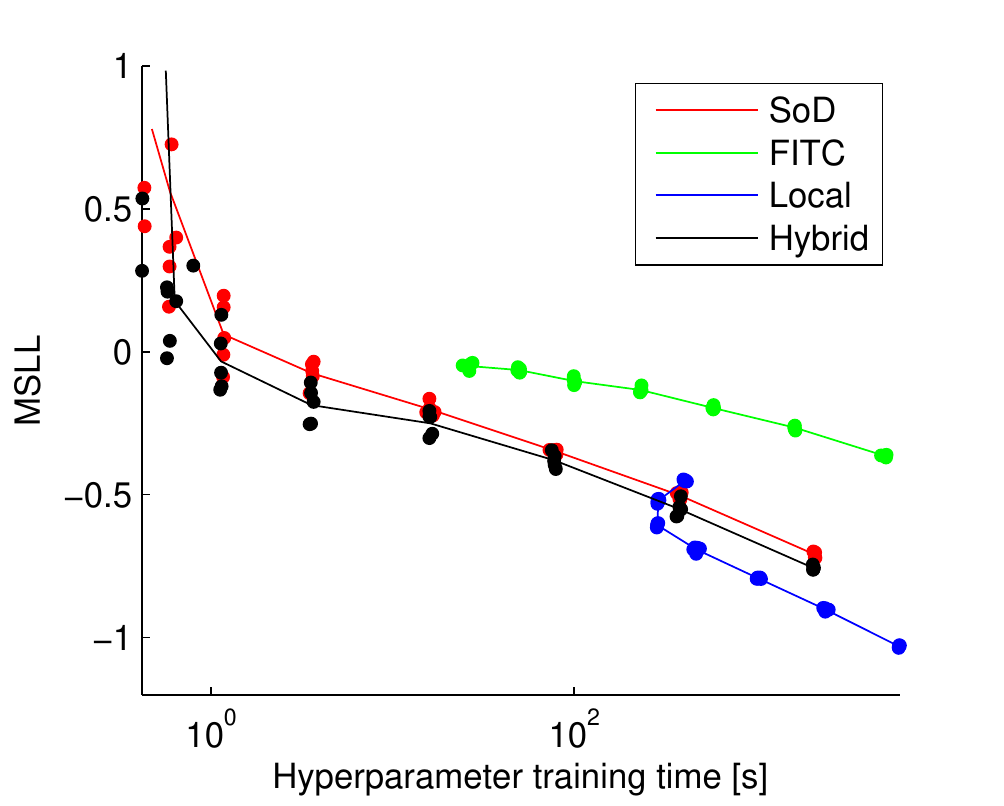}
\label{fig:SYNTH8HypMSLL} &
\hspace*{-0.65cm}\includegraphics[width=5.5cm]{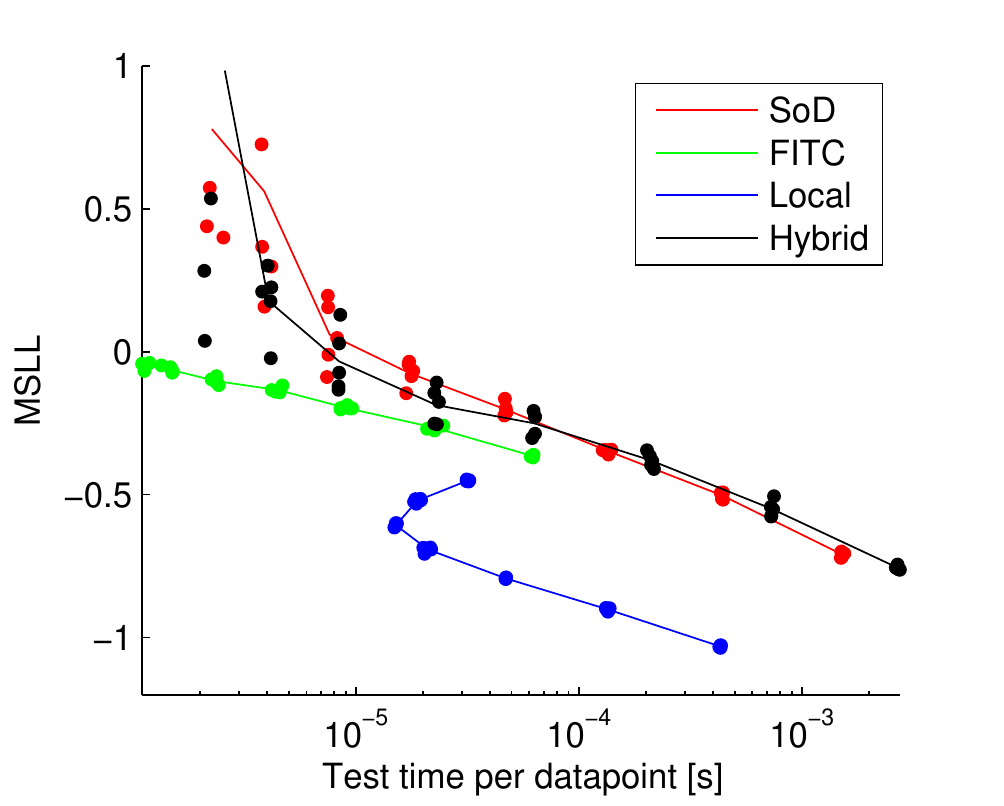}
\label{fig:SYNTH8TestMSLL} \\[-1mm]
\raisebox{4.5cm}{\textsc{chem}} &
\hspace*{-0.42cm}\includegraphics[width=5.5cm]{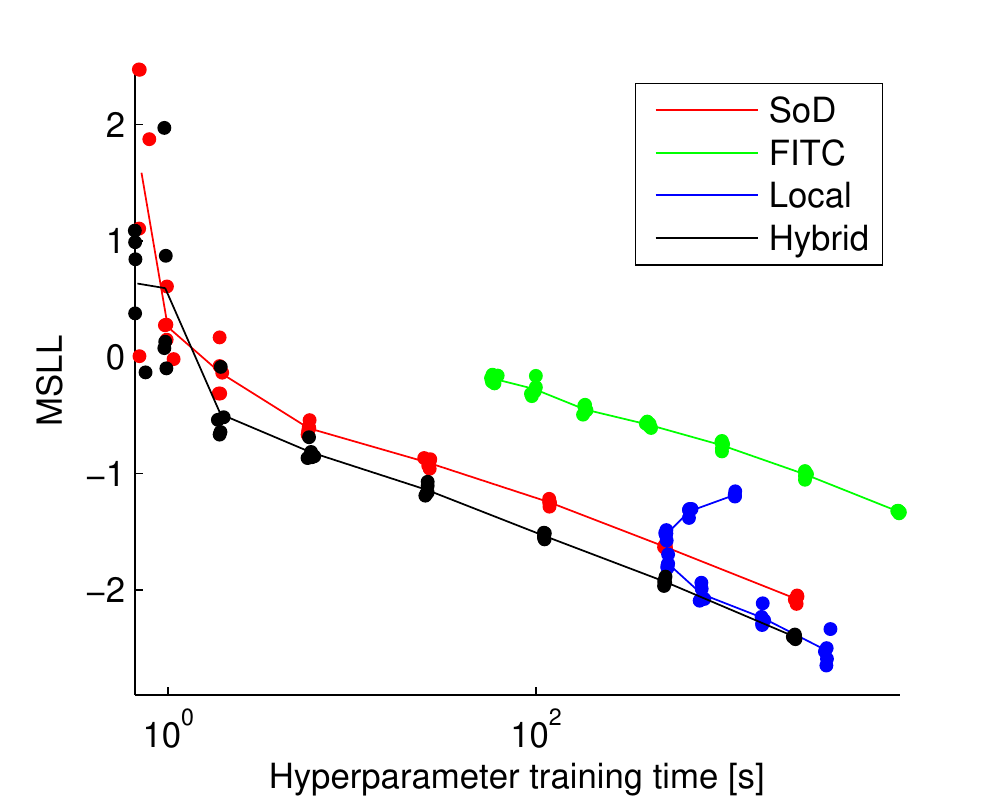}
\label{fig:CHEMHypMSLL} &
\hspace*{-0.41cm}\includegraphics[width=5.5cm]{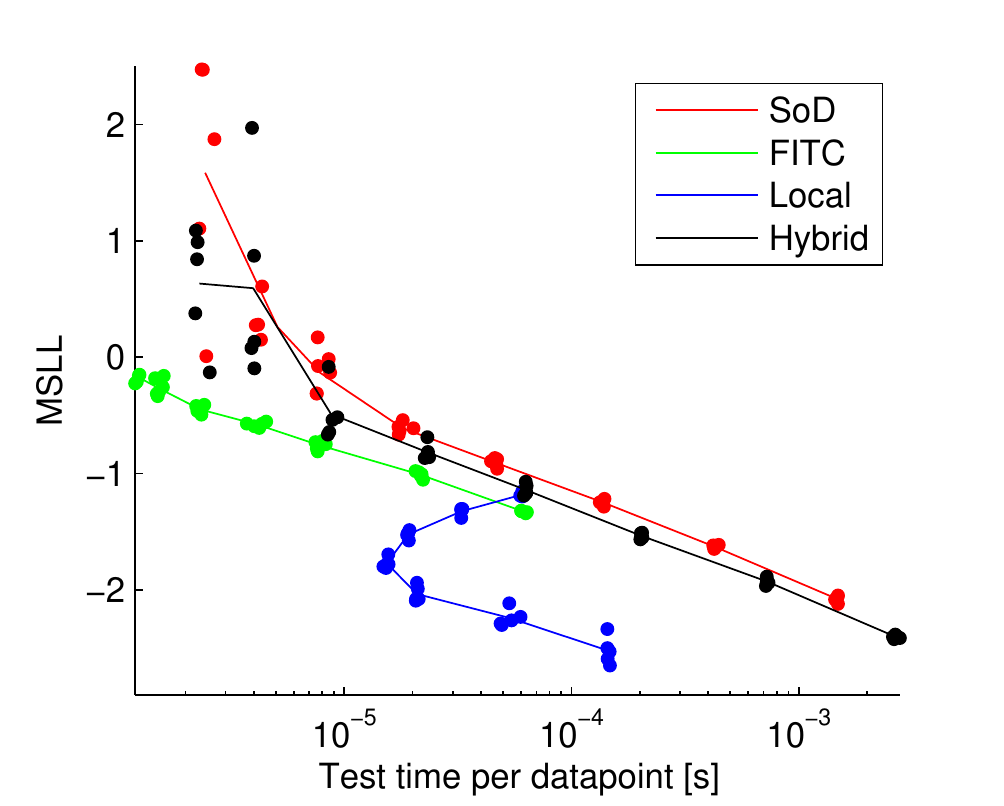}
\label{fig:CHEMTestMSLL} \\[-1mm]
\raisebox{4.5cm}{\textsc{sarcos}} &
\hspace*{-0.42cm}\includegraphics[width=5.5cm]{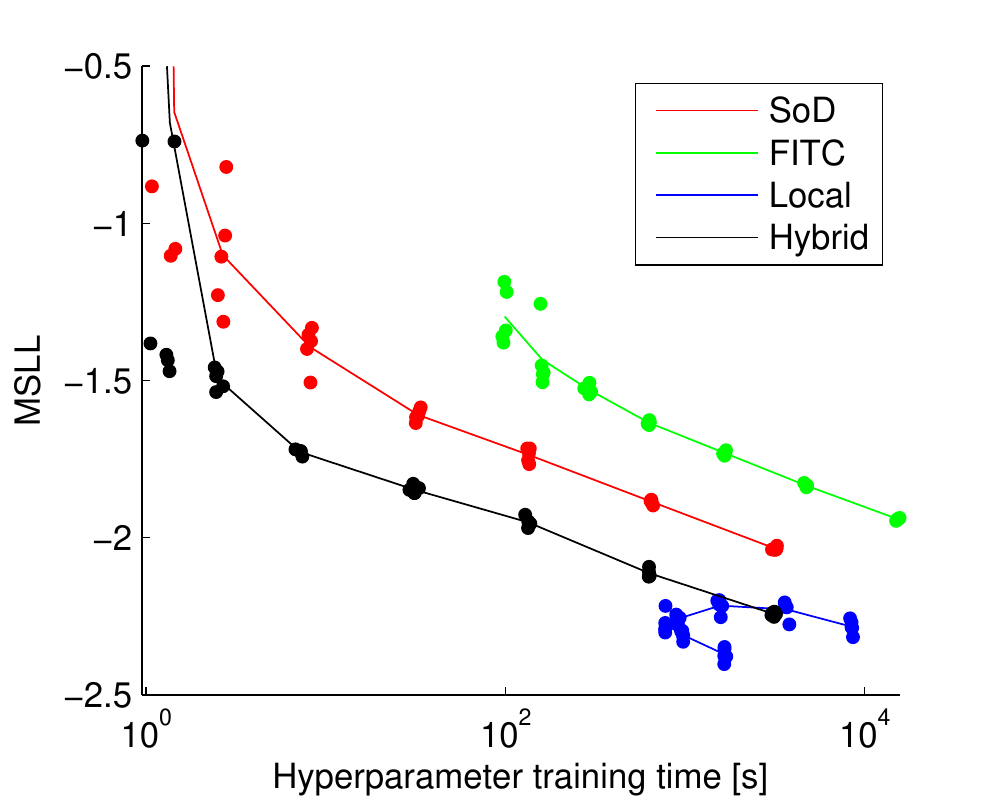}
\label{fig:SARCOSHypMSLL} &
\hspace*{-0.41cm}\includegraphics[width=5.5cm]{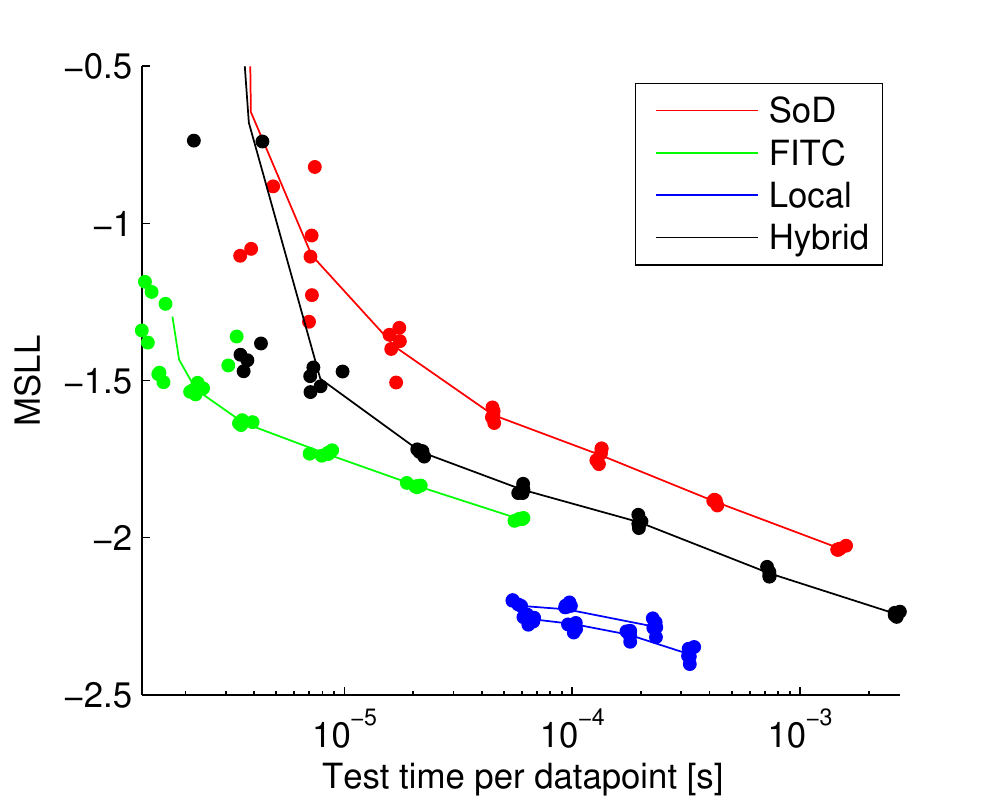}
\label{fig:SARCOSTestMSLL}
\end{tabular}
\end{center}
\caption{MSLL as a function of time (log scale) for the
four datasets. Left: hyperparameter training
time. Right: test time per test point.
Points give the result for each run; lines connect the means of the 5 runs at each $m$.
\label{fig:mslltimeres}}
\end{figure}

In Figure~\ref{fig:timeres} we plot the test set SMSE against
hyperparameter training time (left column), and test time (right column)
for the four methods on the four datasets. Figure
\ref{fig:mslltimeres} shows similar plots for the test set MSLL\@.
When there are further choices
to be made (e.g.\ subset selection methods, joint/separate
estimation of hyperparameters), we generally present the best results
obtained by the method; these choices are detailed at the
end of this section for each dataset individually.
Further details including tables of learned
hyperparameters can be found in \citet{chalupka-11}, although
the experiments were re-run for this paper, so there
are some differences between the two.

The empirical times deviate from theory (Table~\ref{t:approxgpr}) most for
the Local method for small $m$. There
is overhead due to the creation of many small matrices
in \matlab, so that (for example) $m=32$ is always slower (on our
four datasets) than $m=64$ and $m=128$. This effect is demonstrated
explicitly in \cite[Fig.~4.1]{chalupka-11}, and accounts for the bending
back observed in the plots for Local. (This is present
on all four datasets, but can be difficult to see in some of the plots.)

Looking at the hyperparameter training plots (left column), it
is noticeable that SoD and FITC reduce monotonically with increasing
time, and that SoD outperforms FITC on all datasets (i.e.\ for the
same amount of time, the SoD performance is better).  On the test time
plots (right column) the pattern between SoD and FITC is reversed,
with FITC being superior. These results are consistent with theoretical scalings
(Table~\ref{t:approxgpr}): at training time FITC has worse scaling, at
test time its scaling is the same\footnote{In fact, careful comparison
  of the test time plots show that FITC takes longer than SoD; this
  constant-factor performance difference is due to an implementation
  detail in \texttt{gpml}, which represents the FITC and SoD predictors
  differently, although they could be manipulated into the same form.}%
, and it turns out that its more
sophisticated approximation does give better results.

Comparing Hybrid to SoD for hyperparameter learning, we note a
  general improvement in performance for very similar time; this is
  because the additional cost of one FITC training step at the end is
  small relative to the time taken to optimize the hyperparameters
  using the SoD approximation of the marginal likelihood. At test time
  the Hybrid results are inferior to FITC for the same $m$ as
  expected, but the faster hyperparameter learning time means that
  larger subset sizes can be used with Hybrid.

For Local, the most noticeable pattern is that the
run time does not change monotonically with $m$. We also note
that for small $m$ the other methods can make faster approximations
than Local can for any value of $m$. For Local there is
a general trend for larger $m$ to produce better results,
although on \textsc{sarcos} the error actually increases with $m$,
and for \textsc{synth2} the SMSE error rises for
$m=1024, \; 2048$. However, Local often gives
better performance than the other methods in the time regimes where
it operates.

\vspace*{3mm}

We now comment on the specific datasets:

\textsc{synth2}: This function was fairly easy to learn and all
methods were able to obtain good performance (with SMSE close to the
noise level of $10^{-6}$) for sufficiently large $m$.  For SoD and
FITC, it turned out that FPC gave significantly better results than
random subset selection. FPC distributes the inducing points in a more
regular fashion in the space, instead of having multiple close by in
regions of high density. For Local, the joint estimation of
hyperparameters was found to be significantly better than separate;
this result makes sense as the target function is actually drawn from
a single GP\@. For FITC and Hybrid the plots are cut off at
$m=128$ and $m=256$ respectively, as numerical instabilities in the
\texttt{gpml} FITC code for larger $m$ values gave larger errors.

\textsc{synth8}: This function was difficult for all methods
to learn, notice the slow decrease in error as a function of time.
The SMSE obtained is far above the noise level of $10^{-3}$.
Both SoD and FITC did slightly better when selecting the inducing points
randomly. For the Local method, again
joint estimation of hyperparameters was found to be superior,
as for \textsc{synth2}. For both \textsc{synth2} and
\textsc{synth8} we note that the lengthscales learned
by the FITC approximation did not converge to the true
values even for the largest $m$, while convergence was
observed for SoD and Local; see Appendix 1 in
\citet{chalupka-11} for full details.

\textsc{chem}:
Both SoD and FITC did slightly better when selecting the inducing
points randomly. Local with joint and separate hyperparameter
training gave similar results. We report results on the joint method,
for consistency with the other datasets.

\textsc{sarcos}:
For SoD and FITC, FPC gave very slightly better results
than random. Local with joint hyperparameter training did better than
separate training.

\subsection{Comparison with Prediction using the Generative
  Hyperparameters \label{sec:compare_gen}}
For the \textsc{synth2} and \textsc{synth8} datasets it is possible to
compare the results with learned hyperparameters against those
obtained with hyperparameters fixed to the true generative values. We
refer to these as the learned and fixed hyperparameter settings.

For the SoD and Local methods there is good agreement between the
learned and fixed settings, although for SoD the learned setting
generally performs worse on both SMSE and MSLL for small $m$, as would
be expected given the small data sizes. The learned and fixed settings
are noticeably different for SoD for $m \le 128$ on \textsc{synth2},
and $m \le 512$ on \textsc{synth8}.

For FITC there is also good agreement between the learned and fixed
settings, although on \textsc{synth8} we observed that the
learned model slightly outperformed the fixed model by around
0.05 nats for MSLL, and by up to 0.05 for SMSE\@. This may suggest
that for FITC the hyperparameters that produce optimal performance
may not be the generative ones.

\section{Future directions}
We have seen that Local GPR can sometimes make better predictions than
the other methods for some ranges of available computer time. However,
our implementation suffers from unusual scaling behaviour at small~$m$
due to the book-keeping overhead required to keep track of thousands
of small matrices. More careful, lower-level programming than our
\matlab\ code might reduce these problems.

It is possible to combine the SoD with other methods. As a dataset's
size tends to infinity, SoD (with random selection) will always beat
the other approximations that we have considered, as SoD is the only method with no
$n$-dependence (Table~\ref{t:approxgpr}). Of course the other
approximate methods, such as FITC, could also be run on a subset.
Investigating how to simultaneously choose the dataset size to
consider, $n$, and the control parameter of an approximation, $m$, has
received no attention in the literature to our knowledge.

Some methods will have more choices than a single control
parameter $m$. For example, \citet{snelson2006} optimized the
locations of the $m$ inducing points, potentially improving test-time
performance at the expense of a longer training time. A potential
future area of research is working out how to intelligently balance
the computer time spent on selecting and moving inducing points, while
performing hyperparameter training, and choosing a subset size.
Developing methods that work well in a wide variety of contexts without
tweaking might be challenging, but success could be
measured using the framework of this paper.

\cut{
Above we have argued that a natural way to test these
algorithms is to set the hyperparameters using the same
approximation as will be used for training and testing.
However, it would be possible to ``mix and match'' methods;
for example SoD is faster for hyperparameter learning,
but FITC is faster for testing. However, the learned hyperparameters
may be different in these two cases, and it would be interesting
to see how effective a ``mix and match'' strategy would be.
}

\section{Conclusions}
We have advocated the comparison of GPR approximation methods on
the basis of prediction quality obtained vs compute time.  We have
explored the times required for the hyperparameter learning, training
and testing phases, and also addressed other factors that are relevant
for comparing approximations. We believe that future evaluations of GP
approximations should consider these factors (Sec.~\ref{sec:compare}),
and compare error-time curves with standard approximations such as SoD
and FITC\@. To this end we have made our data and code available
to facilitate comparisons.
Most papers that have proposed GP approximations have not
compared to SoD, and on trying the methods it is often difficult to
get appreciably below SoD's error-time curve for the learning phase.
Yet these methods are often more difficult to run and more limited in
applicability than SoD\@.

On the datasets we considered, SoD and Hybrid dominate FITC in terms of
hyperparameter learning. However, FITC (for as long as we ran it) gave
better accuracy for a given test time. SoD, Hybrid and FITC behaved
monotonically with subset/inducing-set size~$m$, making $m$ a useful
control parameter. The Local method produces more varied results, but
can provide a win for some problems and cluster sizes.
Comparison of the iterative methods, CG and DD, to SoD revealed
that they shouldn't be run for a small fixed number of iterations, and
that performance can be comparable with simpler, more stable approaches.
Faster MVM methods might make iterative methods more compelling,
although the IFGT method
only provided a speedup on the \textsc{synth2} problem out
of our datasets. Assuming that hyperparameter learning is the
dominant factor in computation time, the results presented above
point to the very simple Subset of Data method (or the Hybrid variant)
as being the leading contender. We hope this will act as a rallying cry
to those working on GP approximations to beat this ``dumb'' method.
This can be addressed both by empirical evaluations (as presented
here), and by theoretical work.

Many approximate methods require choosing subsets of partitions of
the data. Although farthest point clustering (FPC) improved SoD and
FITC on the low-dimensional (easiest) problem, simple random subset
selection worked similarly or better on all other datasets. Random selection also
has better scaling (no $n$-dependence) for the largest-scale problems.
The choice of partitioning scheme was important for Local regression:
Our preliminary experiments showed that performance was severely
hampered by many small clusters produced by FPC; we recommend our
recursive partitioning scheme (RPC)\@.

\subsubsection*{Acknowledgments}
We thank the anonymous referees whose comments helped improve the
paper.  We also thank Carl Rasmussen, Ed Snelson and Joaquin
Qui\~{n}inero-Candela for many discussions on the comparison of GP
approximation methods.

This work is supported in part by the IST Programme of the European
Community, under the PASCAL2 Network of Excellence, IST-2007-216886.
This publication only reflects the authors' views.

\vskip 0.2in
\bibliography{approxgp}

\end{document}